\PassOptionsToPackage{unicode}{hyperref}
\PassOptionsToPackage{hyphens}{url}
\PassOptionsToPackage{table,dvipsnames,svgnames,x11names}{xcolor}

\documentclass[
letterpaper,
DIV=11,
numbers=noendperiod
]{scrartcl}

\usepackage{cite}
\makeatletter

\renewcommand\@cite[2]{\textsuperscript{\textcolor{blue}{#1\if@tempswa , #2\fi}}}
\makeatother

\usepackage{lineno}
\usepackage{amsmath,amssymb}
\usepackage{iftex}
\ifPDFTeX
\usepackage[T1]{fontenc}
\usepackage[utf8]{inputenc}
\usepackage{textcomp}
\else
\usepackage{unicode-math}
\defaultfontfeatures{Scale=MatchLowercase}
\defaultfontfeatures[\rmfamily]{Ligatures=TeX,Scale=1}
\fi

\usepackage{lmodern}
\IfFileExists{upquote.sty}{\usepackage{upquote}}{}
\IfFileExists{microtype.sty}{
	\usepackage[]{microtype}
	\UseMicrotypeSet[protrusion]{basicmath}
}{}

\makeatletter
\@ifundefined{KOMAClassName}{
	\IfFileExists{parskip.sty}{
		\usepackage{parskip}
	}{
		\setlength{\parindent}{0pt}
		\setlength{\parskip}{6pt plus 2pt minus 1pt}
	}
}{
	\KOMAoptions{parskip=half}
}
\makeatother

\makeatletter
\ifx\paragraph\undefined\else
\let\oldparagraph\paragraph
\renewcommand{\paragraph}{
	\@ifstar
	\xxxParagraphStar
	\xxxParagraphNoStar
}
\newcommand{\xxxParagraphStar}[1]{\oldparagraph*{#1}\mbox{}}
\newcommand{\xxxParagraphNoStar}[1]{\oldparagraph{#1}\mbox{}}
\fi
\ifx\subparagraph\undefined\else
\let\oldsubparagraph\subparagraph
\renewcommand{\subparagraph}{
	\@ifstar
	\xxxSubParagraphStar
	\xxxSubParagraphNoStar
}
\newcommand{\xxxSubParagraphStar}[1]{\oldsubparagraph*{#1}\mbox{}}
\newcommand{\xxxSubParagraphNoStar}[1]{\oldsubparagraph{#1}\mbox{}}
\fi
\makeatother

\usepackage{longtable,booktabs,array}
\usepackage{adjustbox}
\usepackage{calc}
\usepackage{etoolbox}
\usepackage{hhline}
\makeatletter
\patchcmd\longtable{\par}{\if@noskipsec\mbox{}\fi\par}{}{}
\makeatother

\IfFileExists{footnotehyper.sty}{\usepackage{footnotehyper}}{\usepackage{footnote}}
\makesavenoteenv{longtable}
\usepackage{graphicx}
\makeatletter
\newsavebox\pandoc@box
\newcommand*\pandocbounded[1]{%
	\sbox\pandoc@box{#1}%
	\Gscale@div\@tempa{\textheight}{\dimexpr\ht\pandoc@box+\dp\pandoc@box\relax}%
	\Gscale@div\@tempb{\linewidth}{\wd\pandoc@box}%
	\ifdim\@tempb\p@<\@tempa\p@\let\@tempa\@tempb\fi
	\ifdim\@tempa\p@<\p@\scalebox{\@tempa}{\usebox\pandoc@box}%
	\else\usebox\pandoc@box
	\fi%
}
\def\fps@figure{htbp}
\makeatother

\usepackage{authblk}  
\usepackage{tabularx}

\KOMAoption{captions}{tableheading}
\usepackage[section]{placeins}

\usepackage{forest}
\usetikzlibrary{shapes.geometric, arrows.meta}
\usepackage{beramono}
\usepackage{float}
\usepackage{caption}
\usepackage{subcaption}
\usepackage{bookmark}
\usepackage{multirow}
\usepackage{makecell}
\usepackage[table]{xcolor}
\usepackage{colortbl}
\definecolor{HeaderBG}{HTML}{E9E8F5}
\newcolumntype{C}{>{\centering\arraybackslash}X}

\IfFileExists{xurl.sty}{\usepackage{xurl}}{}
\title{FLUID: A Fine-Grained Lightweight Urban Signalized-Intersection Dataset of Dense Conflict Trajectories}

\author[1,2]{Yiyang Chen}
\author[1,2]{Zhigang Wu}
\author[1,2]{Guohong Zheng}
\author[1,2,3]{Xuesong Wu}
\author[1,2]{Liwen Xu}
\author[1,2]{Haoyuan Tang}
\author[1,2,3]{Zhaocheng He\thanks{Corresponding author: hezhch@mail.sysu.edu.cn}}
\author[1,2]{Haipeng Zeng}

\affil[1]{School of Intelligent Systems Engineering, Sun Yat-sen University, Shenzhen, 518107, Guangdong, China}
\affil[2]{Guangdong Provincial Key Laboratory of Intelligent Transportation Systems, Shenzhen, 518107, Guangdong, China}
\affil[3]{Pengcheng Laboratory, Shenzhen, 518055, Guangdong, China}

\date{\vspace{-5ex}}

\begin{document}
\maketitle


\begin{abstract}
The trajectory data of traffic participants (TPs) is a fundamental resource for evaluating traffic conditions and optimizing policies, especially at urban intersections. Although data acquisition using drones is efficient, existing datasets still have limitations in scene representativeness, information richness, and data fidelity. This study introduces \textbf{FLUID}, comprising a fine-grained trajectory dataset that captures dense conflicts at typical urban signalized intersections, and a lightweight, full-pipeline framework for drone-based trajectory processing. FLUID covers three distinct intersection types, with approximately 5 hours of recording time and featuring over 20,000 TPs across 8 categories. Notably, the dataset records an average of 2.8 vehicle conflicts per minute across all scenes, with roughly 15
\% of all recorded motor vehicles directly involved in these conflicts. FLUID provides comprehensive data, including trajectories, traffic signals, maps, and raw videos. Comparison with the DataFromSky platform and ground-truth measurements validates its high spatio-temporal accuracy. Through a detailed classification of motor vehicle conflicts and violations, FLUID reveals a diversity of interactive behaviors, demonstrating its value for human preference mining, traffic behavior modeling, and autonomous driving research.
\end{abstract}

\section*{Background \& Summary}\label{background-summary}
\addcontentsline{toc}{section}{Background \& Summary}

Trajectory datasets of traffic participants (TPs) are fundamental for advancing Intelligent Transportation Systems. Modern data acquisition techniques provide these datasets with unprecedented granularity, enabling the empirical observation of microscopic interactions \cite{jiang2025naturalistic}. Consequently, they support a range of research areas, including microscopic simulation \cite{chen2024data}, behavioral modeling \cite{rowan2025systematic}, and trajectory generation \cite{choi2025gentle}. This level of detail is especially critical at urban intersections, which are dense with complex traffic conflicts. Specifically, the insights gained from this data—covering yielding strategies, violation patterns, and near-collision events \cite{gore2023traffic}—are essential for designing effective control strategies \cite{sarkar2024synergizing}, developing safety interventions, and testing autonomous driving systems. Therefore, high-resolution intersection trajectory data is invaluable for improving urban traffic operations and proactive safety.

Methods for observing traffic participant (TP) mobility and interactions are primarily categorized into ego-centric, roadside, and aerial perspectives \cite{wang2023city}. Ego-centric approaches, foundational to autonomous driving datasets like KITTI \cite{geiger2013vision}, Apolloscape \cite{huang2018apolloscape}, Waymo \cite{sun2020scalability}, and nuPlan \cite{karnchanachari2024towards}, equip vehicles with multi-modal sensors to capture surrounding interactions \cite{wang2023realtime}, but are inherently limited by occlusions and a restricted field of view that prevent a complete scene understanding. Roadside methods, employing either single-sensor (e.g., NGSIM \cite{punzo2011assessment}, Zen Traffic Data \cite{seo2020evaluation}, I-24 MOTION \cite{gloudemans202324}, TJRD \cite{wang2023wide}) or multi-sensor systems (e.g., DLR-UT \cite{schicktanz2025dlr}), offer long-term observation but suffer from limited spatial coverage and potential error accumulation \cite{coifman2017critical}. A critical drawback for both ground-based methods is that visible hardware can alter driver behavior, compromising data naturalness \cite{wang2025comprehensive}. In contrast, aerial observation—typically using drones for their stability and cost-effectiveness—naturally overcomes these limitations by providing a complete, bird's-eye view of the entire traffic scene and all simultaneous interactions within it. Crucially, high-altitude operation ensures the drone remains virtually unnoticed, thus preserving the naturalness of TP behavior. These advantages have spurred the development of numerous drone-based datasets for highways (e.g., highD \cite{krajewski2018highd}, exiD \cite{moers2022exid}, CQSkyEyeX \cite{xu2022cqskyeyex}, MiTra \cite{chaudhari2025mitra}), urban junctions (e.g., inD \cite{bock2020ind}, SIND \cite{xu2022drone}, Hohhot-HDI \cite{pu2025drone}, Songdo Traffic \cite{fonod2025songdo}, RounD \cite{krajewski2020round}), and mixed urban environments (e.g., INTERACTION \cite{zhan2019interaction}, pNEUMA \cite{barmpounakis2020new}, CitySim \cite{zheng2024citysim}).

Among the various traffic scenarios, urban intersections are particularly critical due to the significant efficiency losses and safety risks associated with the interrupted traffic flow. In the domain of intersection datasets, our review focuses on three key aspects: the captured scenes, the provided information, and the quality of trajectory data. Prominent existing datasets typically cover a limited number of scenarios (e.g., 3-4 intersections) \cite{zhan2019interaction, bock2020ind, zheng2024citysim, xu2022drone}, but generally provide comprehensive information, including precise object dimensions and detailed scene elements. In contrast, the HDI dataset \cite{pu2025drone},  while featuring diverse trajectory patterns across more intersections, lacks the aforementioned details. This discrepancy precludes a direct and fair comparison. Therefore, our subsequent analysis evaluates the datasets presented in Table \ref{tab:overall} against these three aspects, with separate considerations for motor vehicles (MVs) and vulnerable road users (VRUs, including pedestrians and two-wheelers).

\begin{table*}[ht!]
	\caption{Summary of intersection drone datasets}
	\label{tab:overall}
	\centering
	\scriptsize                        
	\setlength{\tabcolsep}{3pt}        
	\renewcommand{\arraystretch}{1.15} 
	
	\resizebox{\textwidth}{!}{
		\begin{tabular}{|
				>{\centering\arraybackslash}m{2.2cm} |
				>{\centering\arraybackslash}m{0.9cm}  |
				>{\centering\arraybackslash}m{1.4cm}  |
				>{\centering\arraybackslash}m{1.2cm}  |
				>{\centering\arraybackslash}m{1.05cm} |
				>{\centering\arraybackslash}m{1.05cm} |
				>{\centering\arraybackslash}m{1.0cm}  |
				>{\centering\arraybackslash}m{1.0cm}  |
				>{\centering\arraybackslash}m{1.0cm}  |
				>{\centering\arraybackslash}m{2.0cm}  |
				>{\centering\arraybackslash}m{1.4cm}  |
				>{\centering\arraybackslash}m{1.1cm}  |
				>{\centering\arraybackslash}m{1.2cm}  |
				>{\centering\arraybackslash}m{1.0cm}  |
				>{\centering\arraybackslash}m{1.0cm}  |
			}
			\hline
			\multirow[c]{3}{*}{Dataset} & \multirow[c]{3}{*}{Year}
			& \multicolumn{6}{c|}{\textbf{SCENE}}
			& \multicolumn{5}{c|}{\textbf{INFORMATION}}
			& \multicolumn{2}{c|}{\textbf{DATA QUALITY}} \\
			\cline{3-15} 
			&
			& \multicolumn{2}{c|}{Intersection Type}
			& \multicolumn{4}{c|}{Traffic Flow}
			& \multicolumn{3}{c|}{Object Info}
			& \multicolumn{2}{c|}{Behavior Info}
			& \multicolumn{2}{c|}{Validation} \\
			\cline{3-15}
			&
			& Geometries and channelization
			& Control methods
			& MV arrival rate ($min^{-1}$)
			& VRU arrival rate ($min^{-1}$)
			& Conflict MV ratio
			& Assoc.\ MV ($\le$10 m)
			& \makecell{Structured\\Map}
			& \makecell{MV\\types}
			& \makecell{VRU\\types}
			& Conflict annot.
			& Intention annot.
			& Speed
			& Position \\
			\hline
			
			INTERACTION & 2019
			& FI, TI
			& U (AWS)
			& 34.29 & 1.82 & 1.80\% & 1.46
			& $\checkmark$
			& car
			& pedestrian / bicycle
			& -- & -- & -- & -- \\
			\hline
			
			inD & 2019
			& FI, TI
			& U (RBL/\allowbreak prio.)
			& 13.59 & 9.11 & 3.72\% & 1.25
			& $\checkmark$
			& car
			& pedestrian, bicycle
			& -- & -- & -- & -- \\
			\hline
			
			pNEUMA & 2020
			& Multiple
			& Unknown
			& 7.08 & 3.61 & -- & --
			& -- 
			& \makecell{Car/Taxi, \\ Medium/Heavy \\ Vehicle.,Bus}
			& Motorcycle
			& -- & -- & -- & -- \\
			\hline
			
			CitySim & 2022
			& FI, TI
			& S (prot./\allowbreak perm.) \allowbreak/ U (prio.)
			& 27.69 & -- & 8.14\% & 1.36
			& $\checkmark$ 
			& car
			& --
			& -- & -- & -- & -- \\
			\hline
			
			SIND & 2022
			& FI
			& S (perm.)
			& 21.41 & 11.82 & 11.78\% & 1.74
			& $\checkmark$ 
			& car, truck, bus
			& pedestrian, bicycle, motorcycle
			& -- & $\checkmark$ & -- & -- \\
			\hline
			
			Songdo Traffic & 2024
			& Multiple
			& S (prot.)
			& 2.03 & 0.06 & 3.12\% & 1.28
			& --
			& car/van, truck, bus
			& motorcycle
			& -- & -- & -- & -- \\
			\hline
			
			\textbf{FLUID (ours)} & \textbf{2025}
			& FI, TI, FIDRT
			& S (prot./\allowbreak perm.)
			& 31.10 & 37.29 & 15.14\% & 1.66
			& $\checkmark$ 
			& car, tricycle, van, truck, bus, trailer
			& pedestrian, moped
			& $\checkmark$ & $\checkmark$ & $\checkmark$ & $\checkmark$ \\
			\hline
		\end{tabular}
	}
	
	\medskip
	\begin{minipage}{\textwidth}
		\scriptsize
		\textbf{Notes:}
	\begin{itemize}
		\item \textbf{INTERACTION dataset/pNEUMA dataset/Songdo Traffic dataset:} We use the publicly available multi-agent prediction version of INTERACTION dataset, which features pre-extracted interaction scenarios. For pNEUMA and Songdo Traffic, they do not have structured road network data, so we summarize the types of intersections by observing their sample images. Additionally, the statistical results of pNEUMA were calculated based on 60 major intersections.
		\item \textbf{FI/TI/FIDRT/Multiple:} Abbreviations for intersection geometry and channelization types, as defined in the text. \textit{Multiple} indicates that these types are included, and it also covers other abnormal intersections.
		\item \textbf{S/U:} Signalized/Unsignalized intersection.
		\item \textbf{RBL/AWS/prio./prot./permi.:} Strategies for conflict directions, including Right-Before-Left (RBL), All-Way-Stop (AWS), Priority-controlled (prio.), Protected (prot.), and Permissive (permi.).
		\item \textbf{Arrival rate:} The average number of new MVs/VRUs arriving per minute during the recording period.
		\item \textbf{MV Conflict ratio:} The percentage of total MVs involved in at least one traffic conflict. A detailed calculation is provided in the Methods section.
		\item \textbf{Assoc. MV ($N_{CMVCP}$):} Associated Motor Vehicles. For each conflict pair (CP), this metric is the total count of other conflict-involved MVs located within a 10-meter radius of \textbf{either of the two vehicles} in the pair at the moment of conflict, explained in Figure \ref{fig:process}.
		\item \textbf{annot.:} Annotation. Indicates whether the corresponding information is explicitly labeled in the dataset.
	\end{itemize}
	\end{minipage}
	
\end{table*}

\begin{figure}[htb]
	\centering
	\includegraphics[width=0.6\textwidth]{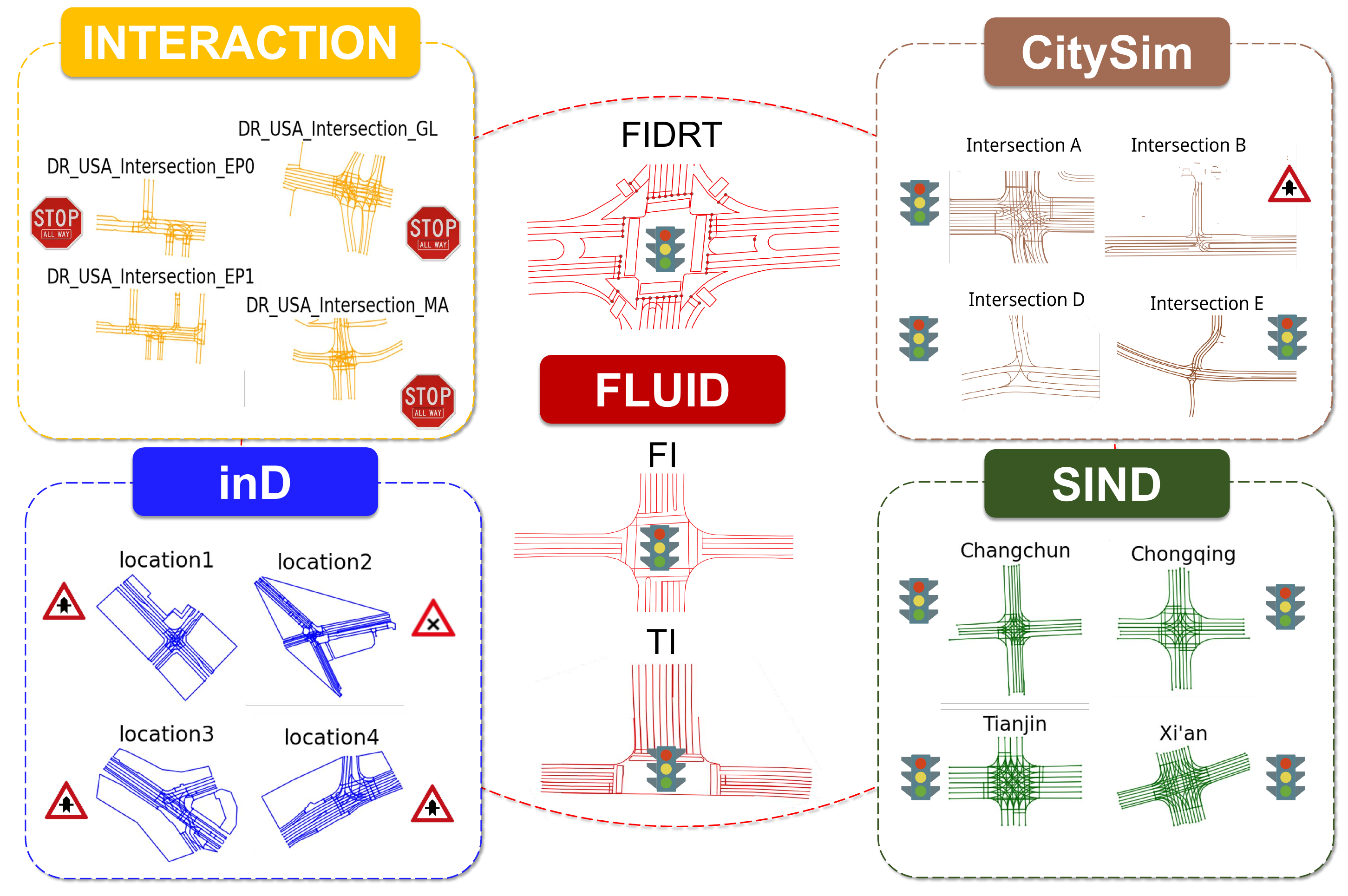}
	\caption{Comparative view of intersection networks from drone-based datasets} 
	\label{fig:int_type}
\end{figure}

The first aspect is the captured scenes, analyzed from both static (intersection type) and dynamic (traffic flow) perspectives.From a static perspective, intersections are classified by their geometries and channelization—such as three-way/four-way intersections (TI/FI), and four-way with dedicated right-turn lanes (FIDRT)—and by their control methods. The latter include unsignalized (e.g., uncontrolled/right-before-left, all-way-stop, priority-controlled) and signalized types, which feature permissive or protected phases for conflicting directions \cite{chandler2013signalized}. Different characteristics of intersections directly affect the speed and angle of TPs when crossing them, which is closely related to the situation of traffic conflicts. However, Figure \ref{fig:int_type} shows that existing drone datasets seldom focus on intersections with dedicated channelization and usually provide incomplete coverage of diverse control strategies. From a dynamic perspective, we evaluate traffic flow using three key metrics: TP arrival rates, MV conflict ratio, and the number of associated MV per conflict ($N_{CMVCP}$), where conflicts are identified using time-based surrogate safety measures (SSMs). A comparative analysis reveals several limitations in existing datasets, such as low or imbalanced participant counts(e.g., a low VRU proportion in INTERACTION and a low overall TP count in inD) and consistently low MV conflict ratio across several major datasets. Furthermore, the pNEUMA dataset’s high shooting altitude leads to decimeter-level target positioning accuracy, making it unsuitable for conflict analysis. In contrast, our proposed FLUID dataset shows clear advantages, featuring a higher overall TP arrival rate and a higher MV conflict ratio of over 15\%. Its $N_{CMVCP}$ metric, comparable to that of SIND, further underscores its strength in capturing dense and interconnected traffic conflicts.

\begin{figure}[htb]
	\centering
	\includegraphics[width=\textwidth]{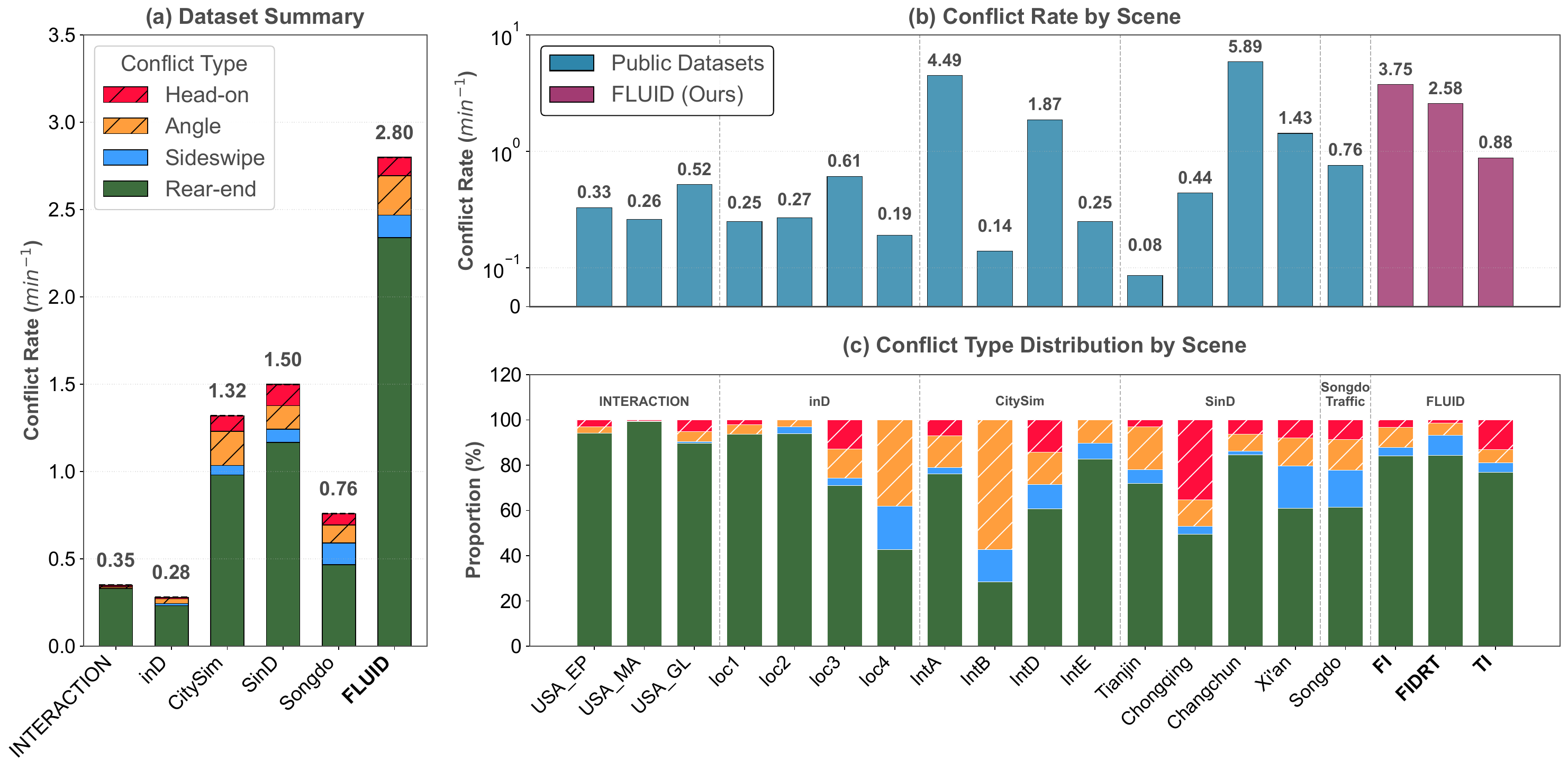}
	\caption{Comparison of conflict types and frequencies across drone-based intersection datasets}
	\label{fig:conflict_rate}
	\begin{minipage}{0.8\textwidth}
		\vspace{0.2cm}
		\footnotesize
		\textit{Notes:} Conflicts are classified by vehicle yaw angle difference; rates shown are conflicts per minute.
	\end{minipage}
\end{figure}

Beyond the scenes themselves, the richness and standardization of the provided information present further challenges. This encompasses both TP attributes and behaviors. From aerial perspectives, the inherent lack of detail hinders both VRU detection and MV classification. Consequently, CitySim omits VRUs entirely, INTERACTION lacks a classification for them, and even in class-annotated datasets like SIND, misclassifications are common. Furthermore, comprehensive behavioral annotations are rare. While analyzing the spatial distribution of traffic conflicts, they seldom provide individual conflict attributes (e.g., object types, conflict types), and the conflicts themselves are often temporally sparse (see Figure \ref{fig:conflict_rate}). Intent information, crucial for understanding driving behavior, is another neglected area, with only the SIND dataset offering preliminary annotations for turning and violation intentions.

The ultimate value hinges on data quality. It presents challenges related to both the final data outputs and the methodological transparency of the generation process. Regarding the final outputs, quantitative assessments of spatio-temporal accuracy are scarce. Most works vaguely mention manual annotation ratios and typically only release polished, final-version trajectories without the corresponding raw data (e.g., videos). This practice prevents users from independently verifying data fidelity or assessing potential error accumulation from over-processing. Methodological transparency is also a widespread issue. Implementation details for detection algorithms are often sparse (e.g., the application of U-Net \cite{bock2020ind}, Mask R-CNN \cite{zheng2024citysim}, YOLOv5 \cite{xu2022drone}) or undisclosed, and descriptions of tracking algorithms are almost universally absent. Furthermore, commercial platforms like DataFromSky \cite{datafromsky-trafficsurvey} or GoodVision \cite{goodvision-traffic-data-collection}are not viable alternatives for many researchers due to barriers such as high costs, accuracy issues, and stringent eligibility requirements. Collectively, the lack of transparency and access to raw data severely undermines the reproducibility of existing datasets.

To address the aforementioned challenges, we introduce FLUID, a new fine-grained trajectory dataset for urban intersections. We conducted 14 flight campaigns at three carefully selected signalized intersections in Xuancheng, Anhui, China, while simultaneously recording their traffic signal states. This process yielded a dataset rich in diverse traffic behaviors, generated via a clear and lightweight pipeline. FLUID is characterized by the following key features:

\begin{itemize}
	\item \textbf{Scene Representativeness}: FLUID features three distinct types of signalized intersections, chosen to cover a range of common traffic conflict types. The high arrival rates of TPs and a significant proportion of conflict-involved vehicles result in a dataset characterized by dense and frequent conflict scenarios.

	\item \textbf{Information Richness}: The dataset provides detailed attributes for multiple classes of TPs. It is further supplemented with synchronized traffic signal states, road maps, and fine-grained annotations of traffic conflicts and behavioral intentions (e.g., turning maneuvers and traffic violations).
	
	\item \textbf{Data Fidelity}: The spatio-temporal accuracy of the trajectory data was validated against the DataFromSky platform and ground-truth measurements from the RTK-GNSS device. Furthermore, we provide a comprehensive description of our entire data acquisition, processing, and fusion pipeline. This provides a basis for assessing the dataset's reliability and extending the methodology to new scenarios.
\end{itemize}

\section*{Methods}\label{methods}
\addcontentsline{toc}{section}{Methods}

\begin{figure}[htb]
	\centering
	\includegraphics[width=0.9\textwidth]{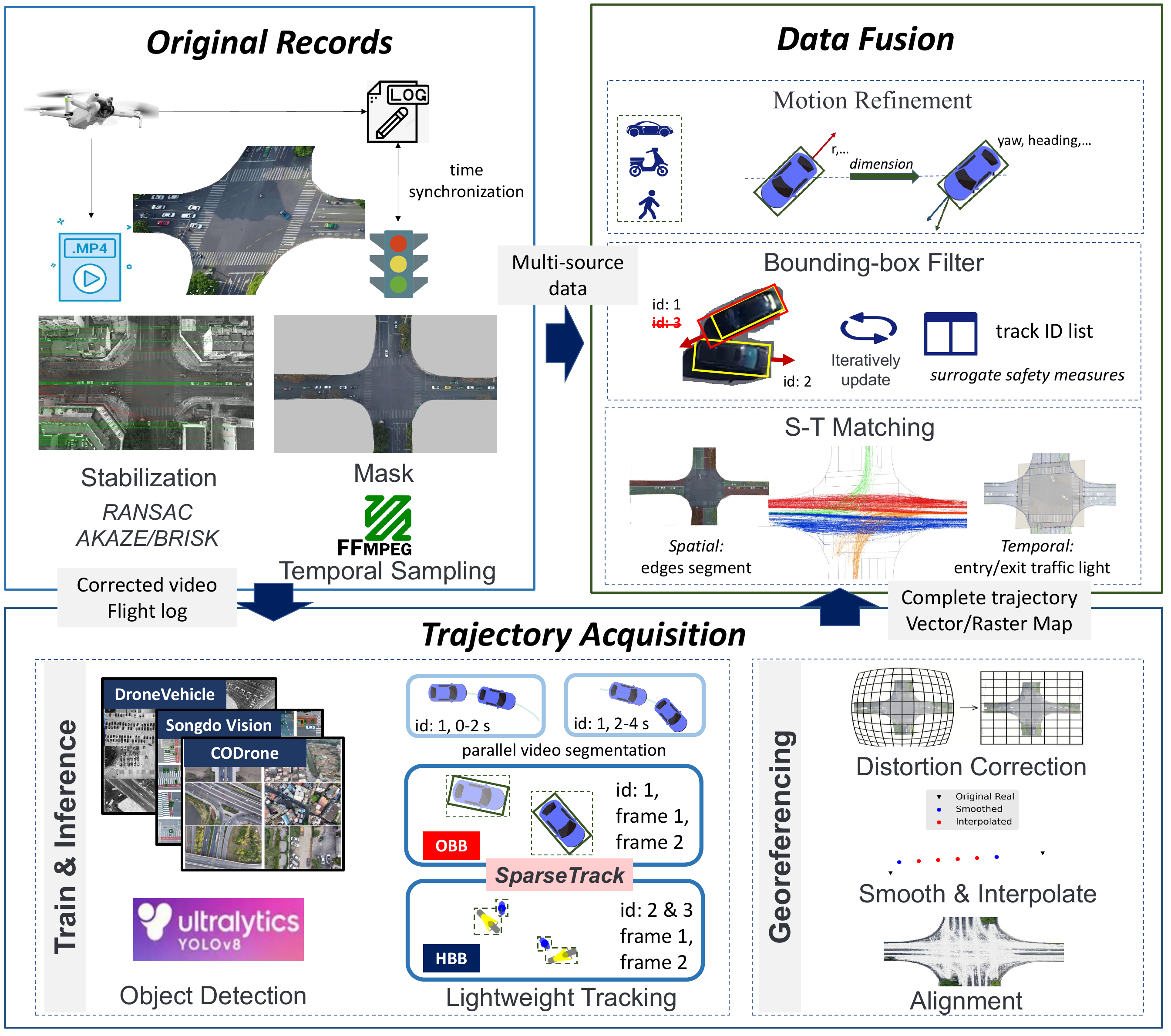}
	\caption{Road map of the construction and quality enhancement of FLUID} 
	\label{fig:framework}
\end{figure}

To obtain high-quality and fine-grained annotated trajectory results from raw collected data, we propose the construction and quality enhancement framework shown in Figure \ref{fig:framework}, which is divided into three parts: raw recording, trajectory acquisition, and data fusion.

\subsection{Original Records}\label{methods:original}

\paragraph{Raw Data.}
The raw data was collected by a three-person team: one operator for the drone, and two observers who recorded traffic signal phases using ground-based cameras. We used a DJI Mini 3 drone to capture high-definition video at 4K resolution (3840×2160 pixels) and approximately 30 \textit{FPS} (29.97 FPS). Due to regulations, the maximum altitude for drones is 120 meters, which is sufficient for capturing microscopic traffic behavior. The drone maintained a consistent flight altitude of $100\sim120$ meters ($100\sim105\; m$ for the FI scenario to capture finer VRU details; $120\;m$ for FIDRT and TI). The plane drift of this type of drone during stationary shooting does not exceed 1.5 meters, with a maximum recording duration per flight of up to 30 minutes. The precise flight altitude for each session can be retrieved from the drone's flight logs. To synchronize the timestamps, the drone and ground cameras were aligned to a unified mobile phone clock with second-level precision, using the coordinated movement of a ground marker as a temporal reference.

\paragraph{Video Pre-processing.}
Due to the drone's lightweight design, the raw footage was susceptible to wind-induced instability and required stabilization. We implemented a two-stage stabilization pipeline based on the open-source tools developed by Fonod et al. \cite{fonod2025advanced}. The lower level employed the feature detector based on video quality: AKAZE \cite{akaze2013} is employed when sharpness remains consistent, leveraging its robust scale-space construction via non-linear diffusion filtering; BRISK \cite{leutenegger2011brisk} is preferred for videos with significant clarity variations. Both offer a favorable balance of speed and accuracy compared to alternatives like SIFT or ORB \cite{tareen2018comparative}. The upper level then utilized a RANSAC algorithm with block matching and motion compensation to robustly estimate inter-frame motion parameters while rejecting false matches from dynamic objects. This process effectively eliminated significant jitter from the footage. Moreover, we develop a masking program to define a Region of Interest (ROI) that strictly encompassed the intersection area. This step served to both anonymize areas outside the road network and reduce the computational load for subsequent processing. The FIDRT scenario was exempted from masking to preserve the full range of VRU activities. Finally, the videos were downsampled to 10 \textit{FPS}. This frame rate is sufficient to capture the decision-making time range of human TPs and is also conducive to detecting the normal motion of VRUs \cite{xu2022drone}. The primary output of this stage is the set of pre-processed videos.

\subsection{Trajectory Acquisition}
The foundation of our analysis lies in the extraction of TP trajectories from videos, which involves identifying track points and associating them with specific individuals in the video.

\paragraph{Object Detection.} 
Detection was accomplished using the YOLOv8 architecture \cite{YOLOv8}, selected for its native support for both horizontal/oriented bounding boxes (HBB/OBB) and its efficiency as a single-stage detector. We prioritized an OBB representation as it provides a tighter and more accurate encapsulation of object boundaries and orientation from our aerial perspective, which is critical for the subsequent analysis of kinematic parameters and TP behaviors. To overcome single-dataset limitations and achieve higher detection accuracy across diverse object classes, we employ a multi-detector ensemble strategy. This strategy involves training the same YOLOv8 architecture independently on three distinct datasets, resulting in three sets of specialised model weights, each optimised to leverage the unique strengths of its training data.

The three specialised detectors were trained on the following datasets:

\begin{itemize}
	\item \textbf{DroneVehicle\_Revised}: Our custom-curated dataset, which augments the DroneVehicle benchmark \cite{sun2022drone} with manually-annotated samples of underrepresented classes (e.g., tricycles from FLUID) and additional vehicle types (e.g., motorcycles from the VETRA dataset \cite{hellekes2024vetra}) to enhance its robustness.
	\item \textbf{CODrone} \cite{ye2025more}: A recent high-quality, \textbf{4K-resolution} OBB detection benchmark, used to ensure state-of-the-art performance on common object categories, especially small objects such as pedestrians and two-wheelers.
	\item \textbf{Songdo Vision} \cite{fonod2025advanced}: A third detector was trained on this dataset to enhance the detection of VRUs (e.g., motorcycles). Although notable for its extensive and precise HBB annotations, its HBB detections were subsequently converted to the OBB format to serve as a supplement to the primary OBB detectors.
\end{itemize}

\begin{figure}[htb]
	\centering
	\includegraphics[width=0.9\textwidth]{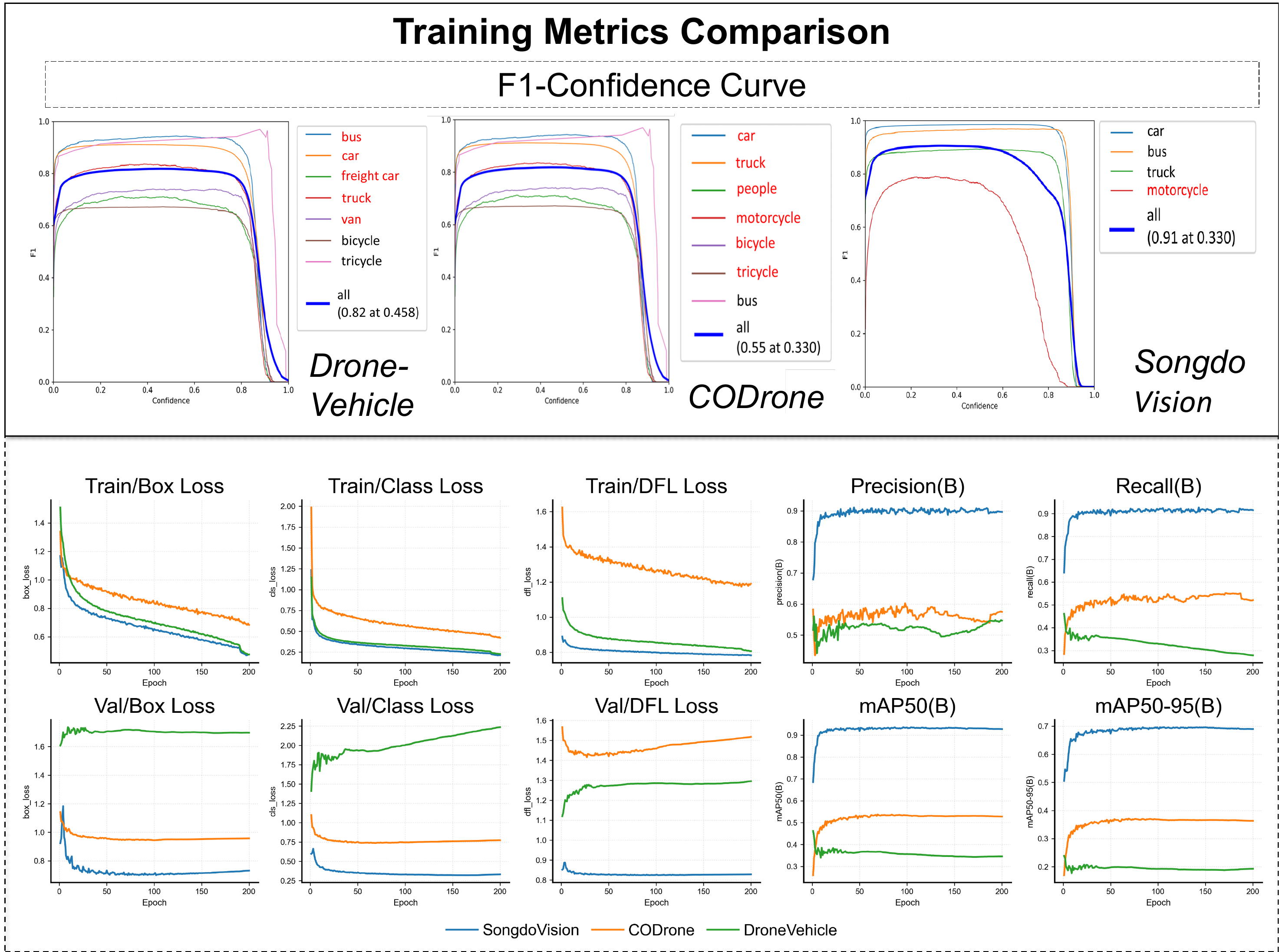}
	\caption{Training result metrics of three detection models (the \textit{advantage categories} of the model are marked in red)} 
	\label{fig:metrics}
\end{figure}

For each of the three training processes, the respective dataset was re-partitioned into training, validation, and test subsets using a 70\%/20\%/10\% ratio, and each model instance was trained for 200 epochs. Subsequently, the final detections were generated through a category-based fusion of the outputs from these specialised detectors, as illustrated in Figure \ref{fig:metrics}.

To evaluate the performance of each detector, we adopt the standard metrics from the YOLOv8 framework, including bounding box loss, classification loss, and Distribution Focal Loss (DFL) for both training and validation phases. The detection performance is quantified by Precision, Recall, and mean Average Precision (mAP) at Intersection over Union (IoU) thresholds of 0.5 and 0.5:0.95. Detailed definitions of each metric are provided in the supplementary materials \cite{zou2023object}\cite{li2020generalized}. F1-Confidence curve is a curve that shows the variation of F1-score as Confidence gradually increases. The F1-score exceeded 0.8 for all categories, indicating a well-balanced performance between precision and recall and confirming the robustness of the detection model.

In this process, We identified the object categories where each detector outperforms the others (termed its \textit{advantage categories}). The final detection output for any given frame was constructed by integrating only the detections for these designated advantage categories from their respective specialist detector. This ensemble approach ensures a comprehensive and precise set of object detections.

\paragraph{Lightweight Tracking}
Once object detection was complete, we employed SparseTrack \cite{liu2025sparsetrack} to link the detections over time and form continuous trajectories. This algorithm was chosen for its effectiveness in dense and complex scenes, as it uses only intersection over union (IoU) for matching. As an enhancement to the widely-used ByteTrack \cite{zhang2022bytetrack}, SparseTrack introduces pseudo-depth estimation and deep cascade matching (DCM). ensuring robustness against occlusions and in mixed-traffic scenarios with VRUs. The pseudo-depth ($d_p$) is defined as:

\begin{equation}
	L_p=H-y_p
\end{equation}

The pseudo-depth of a target is determined by its distance from the camera. This value, denoted as $d_p$ (where a larger value implies a greater distance), is calculated using the image height $H$ and the y-coordinate of the bounding box's bottom-center point, $y_p$, within the image's pixel coordinate system. Next, the Deep Cascade Matching (DCM) algorithm refines the association process for confirmed tracks by assigning matching priorities. This multi-level strategy employs a cost matrix that combines cosine and Mahalanobis distances alongside the standard IoU score. The final data association is then resolved using the Hungarian algorithm.

For efficient inference on large video files, we employed a stream-based reading and parallel preprocessing pipeline to enable lightweight data loading and mitigate the risk of out-of-memory (OOM) errors.

\paragraph{Georeferencing.}
To georeference the pixel-based trajectories, we performed camera calibration and lens distortion correction. Initial intrinsic and extrinsic parameters were sourced from the image metadata and the drone's flight logs. The distortion coefficients were then refined via a least-squares optimization. This process minimized the discrepancy between theoretical distances, calculated using the Ground Sampling Distance (GSD) from a constant flight altitude, and measured pixel distances in the pre-stabilized video feed. These coefficients account for both radial distortion ($k_1, k_2, k_3$) and tangential distortion ($p_1, p_2$):

\begin{equation}
	\mathrm{Dist}=[k_1,k_2,p_1,p_2,k_3]
\end{equation}

Given that the distortion function is modeled as a polynomial, the corrected coordinates $(x_{dist},y_{dist})$ can be calculated from $(x,y)$ and the coefficient $\mathrm{Dist}$, where $(x,y)$ is the normalized coordinate computed from the pixel position using the camera intrinsic parameters.

Following distortion correction, we obtained a set of refined pixel coordinates for each trajectory. These trajectories, however, contained jitter and missing frames resulting from residual stabilization errors, occlusions, or indistinct object features. A multi-stage process was implemented in the pixel domain to refine these trajectories, focusing on interpolation and smoothing: 

\begin{itemize}
	\item Savitzky-Golay (S-G) Filter: The S-G filter with a dynamic window size was applied to the pixel coordinates of the bounding box vertices and their orientation angles. This procedure mitigates high-frequency jitter in the raw detections.
	\item Kinematic Interpolation: Missing data points in the trajectories were interpolated. The positions were estimated by calculating the linear and angular velocities from adjacent valid data points in the pixel coordinate system. This kinematic method was supplemented by linear and nearest-neighbor interpolation as fallback routines. All interpolated points were explicitly flagged.
	\item Rauch-Tung-Striebel (RTS) Smooth: A RTS-smoother was applied to the complete trajectories (containing both original and interpolated points). It is based on a Constant Velocity (CV) motion model. From the resulting smoothed velocity profile, acceleration was computed using the central difference method.
\end{itemize}

\begin{figure}[htb]
	\centering
	\includegraphics[width=0.9\textwidth]{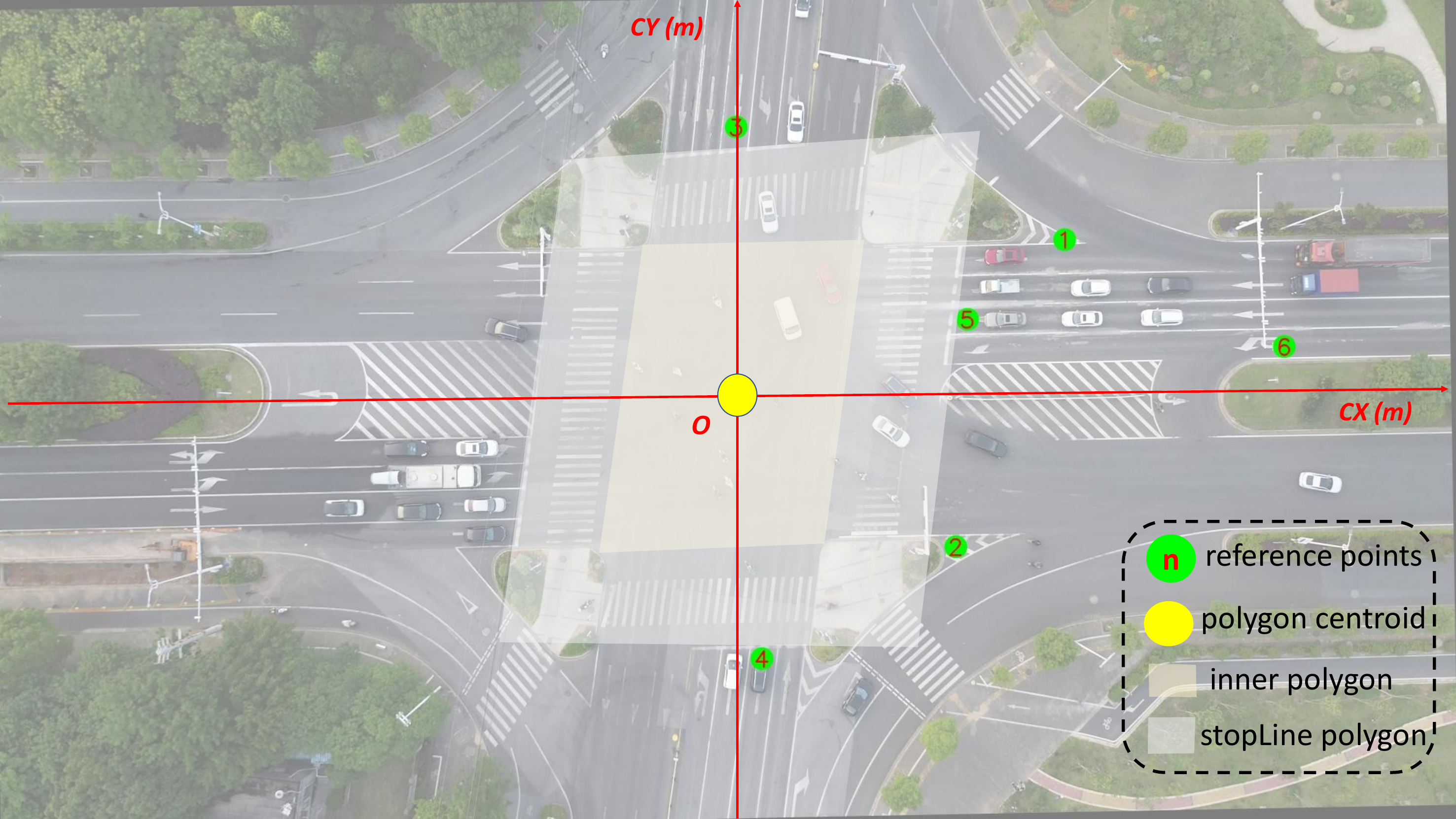}
	\caption{Illustration of local geographic coordinate system construction} 
	\label{fig:coordinate}
\end{figure}

Upon completion of the pixel-domain processing, the trajectories were georeferenced to a local Cartesian coordinate system (in meters), as illustrated in Figure \ref{fig:coordinate}. This was achieved by first computing a homography matrix that mapped pixel coordinates to WGS84 geodetic coordinates. The matrix was calibrated using several ground control points, whose positions were precisely measured with an RTK-GNSS device. These geodetic coordinates were then projected into the final local system via a Universal Transverse Mercator (UTM) projection. The origin of this local system is the centroid of the intersection's inner polygon, which is derived from a larger \textrm{stopLine polygon} by excluding the crosswalk areas. The \textrm{stopLine polygon} itself is the area bounded by the stop lines and is used for subsequent violation analysis.

\subsection{Data Fusion}
This process is applied to complete trajectories, aiming to fuse tracks of diverse TP types from multiple sources and align them through spatio-temporal matching.

\paragraph{Motion Refinement.}
A stable representation for each target's physical dimensions was established by searching the median width and height (converted to meters) from its complete set of detections. Subsequently, we performed kinematic correction by computing and refining two orientation angles: heading (direction of motion) and yaw (TPs' longitudinal orientation). The raw yaw sequence was first stabilized via a bidirectional method, which traverses the sequence to identify stable values and replace intermittent outliers, mitigating abrupt changes. Concurrently, the heading angle, computed from the target's displacement vector, was used as a reference to correct anomalous yaw values that deviated significantly from the direction of motion.

\begin{figure}[htb]
	\centering
	\includegraphics[width=0.9\textwidth]{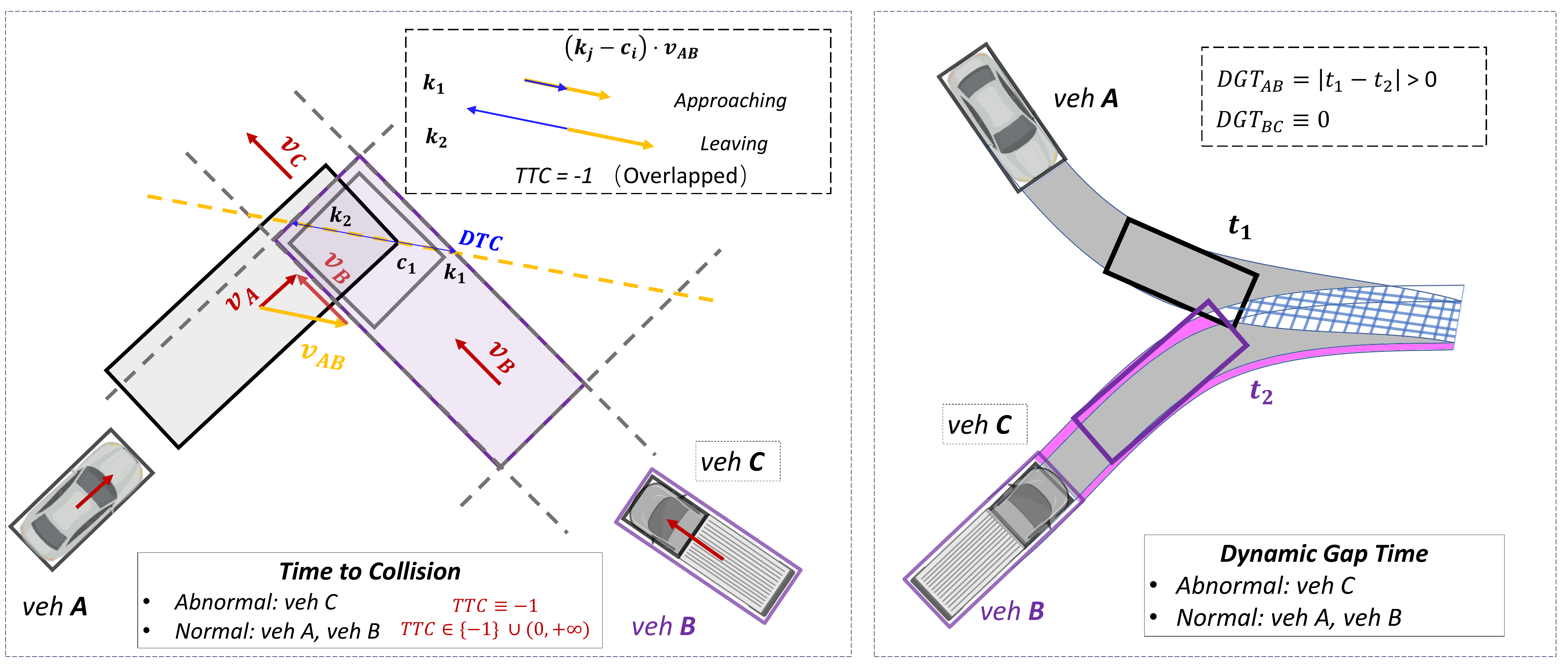}
	\caption{Filtering abnormal vehicles based on 2D-SSMs calculations (left: TTC, right: DGT)} 
	\label{fig:2DSSM}
\end{figure}

\paragraph{Bounding-box Filter.} Erroneous and redundant bounding boxes were then filtered in a two-stage process. First, a heuristic pass removed trajectories with a short duration, minimal displacement, or an average confidence below 0.5, targeting false positives like static objects and shadows. The second stage resolved persistent overlaps on single objects (i.e., dual detections caused by classification ambiguity). Although Surrogate Safety Measures(SSMs) are typically used to analyze traffic conflicts \cite{arun2021systematic}, their inherent ability to quantify spatio-temporal proximity provides a new perspective for lightweight trajectory post-processing. We perform scene-wide removal of redundant bounding boxes using Two-dimensional SSMs (2D-SSMs)—specifically \textbf{Time-to-Collision} (TTC) and \textbf{Dynamic Gap Time} (DGT)—which distinguish persistent redundancy from transient overlaps. As shown in Figure \ref{fig:2DSSM}, vehicle A and B represent valid, distinct targets, whereas vehicle C is abnormal.

As depicted in the left panel of Figure \ref{fig:2DSSM}, we consider a scenario where at least one of two objects, vehicle A and vehicle B, is in motion. Building upon the open-source work of Jiao et al. \cite{jiao2023ttc}, we compute TTC to assist in determining potential overlaps between the bounding boxes of these two relatively moving objects.

Let $v_{AB}$ be the relative velocity vector between vehicle A and vehicle B. For each corner point $c_i$ of vehicle A's bounding box, we define $k_j$ as the intersection point of the line originating from $c_i$ with direction $v_{AB}$ and the line segments forming the bounding box of vehicle B. If such an intersection point $k_j$ exists, the vector $(k_j - c_i)$ represents the relative displacement from the corner point $c_i$ to the edge of vehicle B's bounding box. To determine whether an edge is approaching or receding from a corner point, we compute the scalar product of this displacement vector with the relative velocity vector, $v_{AB}$. As illustrated in the figure, for the intersection point $k_1$, the scalar product $v_{AB} \cdot (k_1 - c_i)$ yields a positive value, indicating that the corresponding edge of B is approaching corner $c_i$. Conversely, for $k_2$, the product $v_{AB} \cdot (k_2 - c_i)$ is negative, signifying that the edge is receding from $c_i$. We use an indicator function $I(c_i)$ to denote this relationship, where $I(c_i) = +1$ for an approaching edge and $I(c_i) = -1$ for a receding one.

Let $d_{i\rightarrow j}$ denote the minimum distance from corner $c_i$ to the bounding box of vehicle B along the direction of the relative velocity vector $v_{AB}$. It is calculated as:

\begin{equation}
	d_{i \rightarrow j}=\left\{\begin{array}{rl}
		\left\|\boldsymbol{k}_{\boldsymbol{j}}-\boldsymbol{c}_i\right\|, & \left(\boldsymbol{k}_j-\boldsymbol{c}_i\right) \boldsymbol{v}_{A B} \geq 0 \\
		\inf, & \left(\boldsymbol{k}_j-\boldsymbol{c}_i\right) \boldsymbol{v}_{A B}<0 \text { or } \boldsymbol{k}_j \text { does not exist }
	\end{array}\right.
\end{equation}

By iterating over all corner points $c_i$ of vehicle A, we define the Distance-to-Collision (DTC) as the minimum magnitude among all possible distances $d_{i\rightarrow j}$. The DTC between vehicles A and B at the current time step is thus given by:

\begin{equation}
	\begin{split}
		DTC_{A\rightarrow B} &= \min{\mathbf{D}}_{i\rightarrow j}\\
		DTC &= \min\{DTC_{A\rightarrow B},DTC_{B\rightarrow A}\}
	\end{split}
\end{equation}

The, TTC can be calculated from the DTC as follows:

\begin{equation}
	\mathrm{TTC} = \begin{cases} -1, & \text{if } \left(\sum_{c \in C_i} I_+(c) > 0\right) \land \left(\sum_{c \in C_i} I_-(c) > 0\right) \\ \dfrac{\min\{\|k_c - c\| \cdot {I_+(c)}}{\|v_{ij}\|}, & \text{if } \sum_{c \in C_i} I_+(c) > 0 \text{ and } \sum_{c \in C_i} I_-(c) = 0 \\ \infty, & \text{if } \sum_{c \in C_i} I_+(c) = 0 \end{cases}
\end{equation}

The decision rule for TTC is as follows: a value of -1, which indicates that parts of the two objects are simultaneously approaching and receding, signifies a bounding box overlap. In such cases, the trajectory with the shorter duration is identified as redundant and is removed.

However, the TTC-based method is insufficient for cases where an erroneous bounding box moves in close parallel with the true target, maintaining a near-constant relative velocity (i.e., they are relatively static). To address this limitation, we introduce DGT, as illustrated in the right panel of Figure \ref{fig:2DSSM}.

The most common time-based SSM, Post-Encroachment Time (PET), is ill-suited for this task because it requires a clear \textit{exit time} from a conflict zone, which is ambiguous when considering full bounding boxes. We therefore turned to the concept of Gap Time (GT), which only considers the time difference between the two vehicles entering the conflict zone \cite{wang2021review}. Based on this, we define DGT. We generalize the conflict point to a conflict area—the intersection of the two bounding boxes' swept areas, detected via the Separating Axis Theorem (SAT)—and calculate DGT as the time difference between the moments each vehicle first enters this area. A DGT that remains zero for a sustained duration signifies a persistent overlap, prompting the removal of the shorter trajectory.

By applying this sequential filtering process (first TTC, then DGT), we effectively screened out the majority of overlapping and erroneous bounding boxes, retaining only the validated TP trajectories.

\paragraph{S-T Matching.}
The final step in our data fusion pipeline is the precise spatio-temporal matching of the validated TP trajectories.

Spatial matching was achieved by integrating the trajectory data with geographic information. For each aerial video, a georeferenced TIFF image with a local Cartesian coordinate system was generated. Then, the intersection's road layout was semantically segmented. We partitioned each road edge into its constituent entry and exit lane groups. By jointly matching a trajectory's position and its refined yaw angle against these directional layers, we were able to assign a specific turning movement (e.g., left-turn, straight, right-turn) to each TP.

In the temporal matching stage, we focused on time-series data. Each trajectory was synchronized with intersection entry/exit time and the corresponding traffic signal status. This temporal integration allowed for the fusion of supplementary information, such as traffic violations (e.g., red-light running).

\section*{Data Records}\label{data-records}
\addcontentsline{toc}{section}{Data Records}

The full dataset is available through the figshare repository \cite{Chen2025}. Figure \ref{fig:scene} shows all the intersections, which are located in the central urban area. For privacy protection, the specific names and locations of these three intersections have been anonymized. Data was collected at these sites during clear daytime hours on selected days in January and May 2025. After an anomaly screening process, a total of approximately 5 hours of raw video footage was obtained. An overview of these three intersections is as follows:

\begin{figure}[htb]
	\centering
	\includegraphics[width=\textwidth]{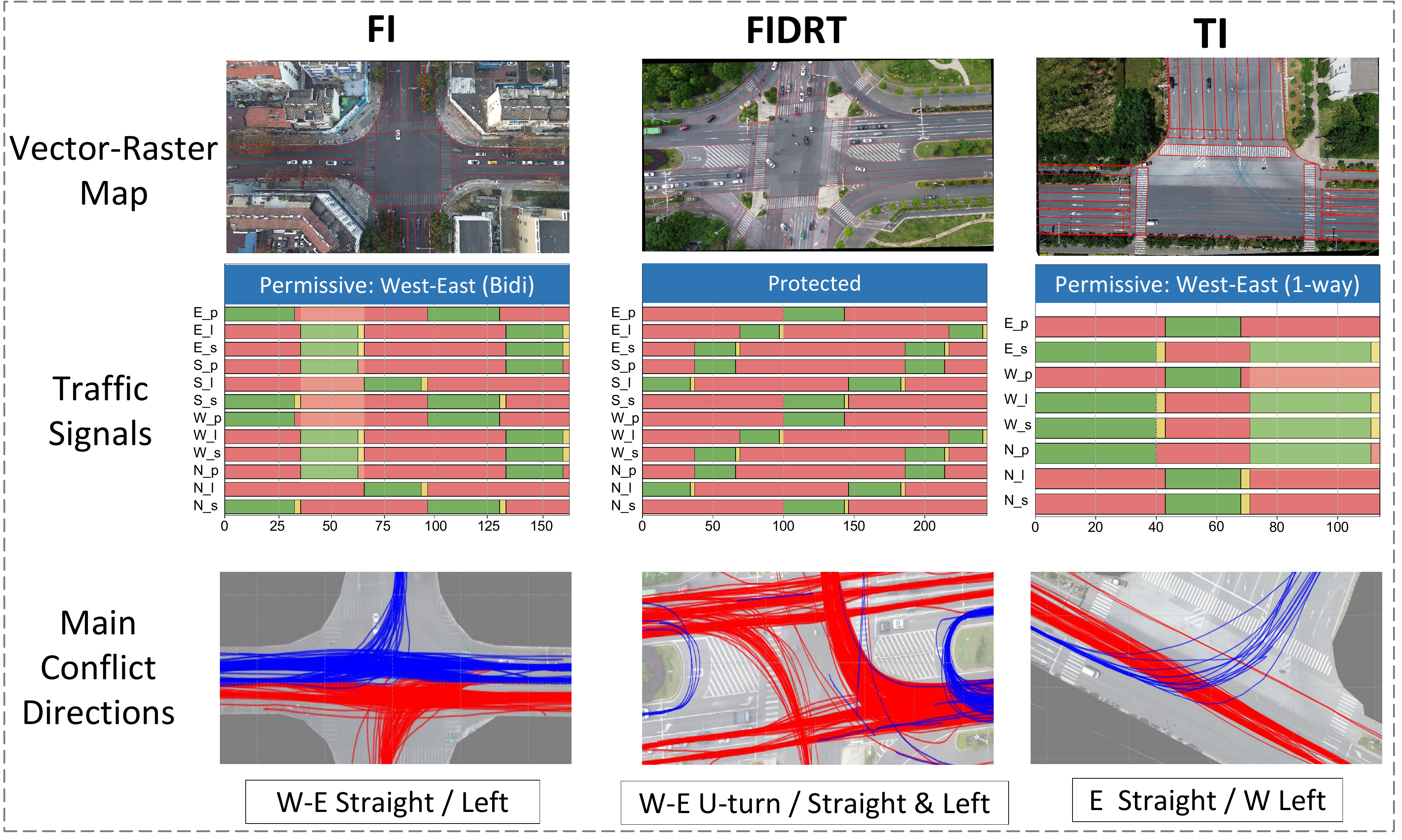}
	\caption{Scene information recorded in FLUID} 
	\label{fig:scene}
\end{figure}

\begin{enumerate}
	\item \textbf{FI} (Four-way Intersection): This intersection is governed by a three-phase signal control. It is characterized by high volumes of MVs and VRUs, leading to direct conflicts. Furthermore, conflicts arise from the concurrent release of through and left-turning traffic in the east-west direction.
	\item \textbf{FIDRT} (Four-way Intersection with Dedicated Right-turn Lanes): Operating on a four-phase signal, this intersection has no direct internal conflict points but exhibits high traffic density. Conflicts are present externally at U-turn spots where paths cross with other traffic streams.
	\item \textbf{TI} (T-Intersection): This three-way intersection is managed by a two-phase signal. It serves as a valuable site for observing frequent conflicts between opposing through and left-turning vehicles, particularly in lower-volume traffic scenarios.
\end{enumerate}

\begin{figure}[htb]
	\centering
	\includegraphics[width=\textwidth]{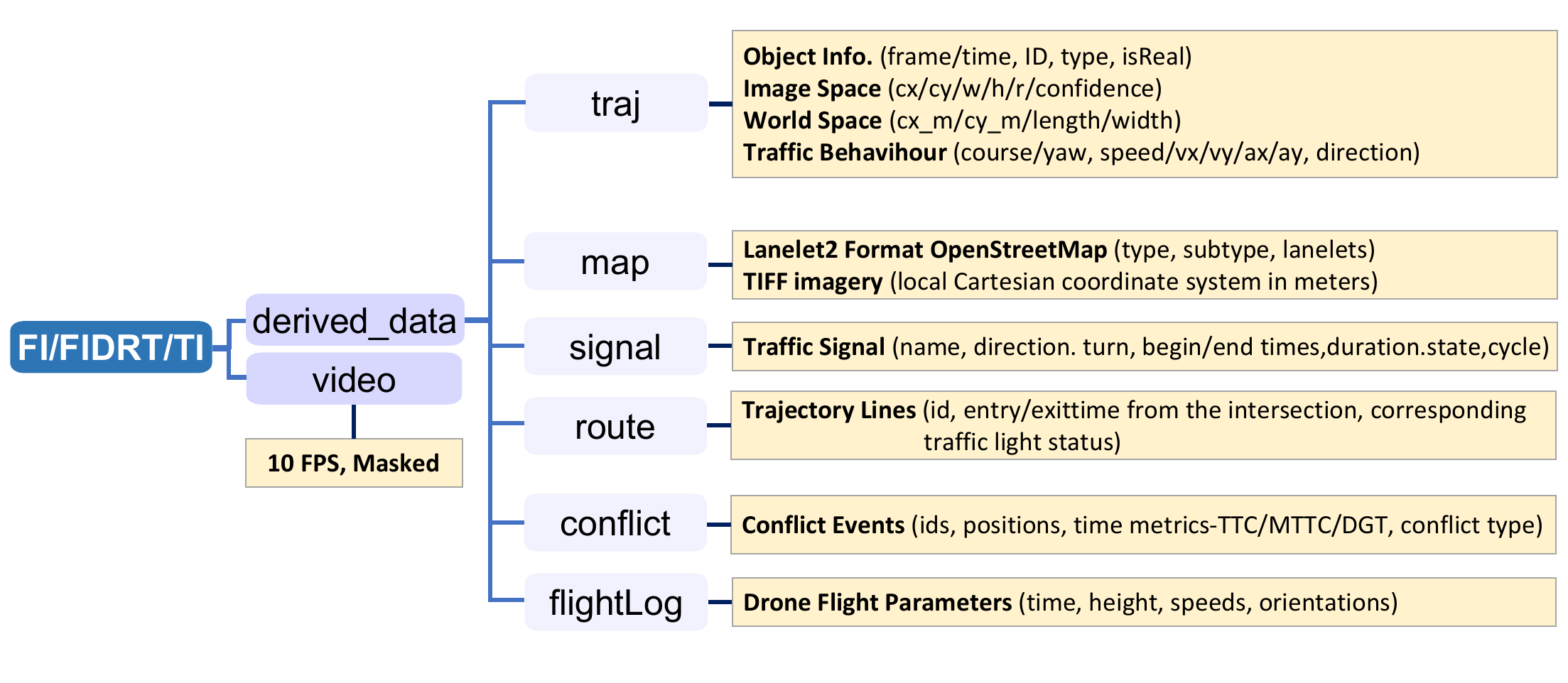}
	\caption{Structure of files of FLUID}
	\label{fig:dataset_structure}
\end{figure}

The formats of files in the FLUID dataset are: MPEG-4 Part 14 (\textbf{MP4}), Comma-Separated Values (\textbf{CSV}), Tagged Image File Format (\textbf{TIFF}), and OpenStreetMap (\textbf{OSM}). Since there are too many fields, their meanings are explained in the Markdown document \textbf{README.md}. As the structure shown in Figure \ref{fig:dataset_structure}, the following components are available:

\begin{itemize}
\item Privacy-preserved videos (\textit{video}): Provided in MP4 format at 10 \textit{FPS}.
\item Signal timings (\textit{signal}): Manually annotated signal control data in CSV format, sampled per second.
\item Flight log during videos (\textit{flightLog}): Drone flight posture recording in CSV format.
\item Maps (\textit{map}): Georeferenced TIFF images and vector maps largely compatible with the Lanelet2 (OSM) format \cite{poggenhans2018lanelet2}.
\item Processed trajectories (\textit{traj/route/conflict}): Offered in CSV format (the annotations for conflicts and violations belong to different files than the original tracks).
\end{itemize}

It is worth noting that FLUID offers not only a unified processing framework, but also high-quality trajectory extraction results, which act as a solid baseline and can be further enhanced by the research community.

\section*{Technical Validation}\label{technical-validation}
\addcontentsline{toc}{section}{Technical Validation}

The validation of the FLUID dataset is \textbf{three-fold}. First, we assess the effectiveness of our data processing pipeline by benchmarking its results against the DataFromSky platform and ground truth data. Second, we establish the significance of our chosen scenes by contrasting their conflict profiles with those of other public datasets. Finally, we confirm the dataset's richness by demonstrating that each of the three scenes features a unique distribution of conflicts and violations, capturing a wide spectrum of behaviors.

\subsection{Trajectory Accuracy}

Previous datasets have rarely benchmarked their results against alternative processing methods or conducted systematic accuracy validation using supplementary data sources. Acknowledging that many prominent trajectory datasets—such as pNEUMA \cite{barmpounakis2020new}, MAGIC \cite{ma2022magic}, and Mitra \cite{chaudhari2025mitra}—rely on the DataFromSky (DFS) \cite{datafromsky-trafficsurvey} platform, we selected DFS as a benchmark to validate our FLUID framework's effectiveness. Furthermore, inspired by vehicle kinematics studies that often leverage ground-based positioning for validation \cite{mi2024capturing}, we equipped a test vehicle with an RTS-GNSS device to establish a high-precision ground truth trajectory.

For this validation, we collected ground control points for georeferencing, as shown in Figure \ref{fig:coordinate}. We utilized the 5-minute video analysis offered by the DFS free tier, processing the first five minutes of footage from the FIDRT scene recorded on May 26, 2025. This DFS-generated trajectory set serves as a baseline. We then compared it against the trajectories extracted by our FLUID framework from two video sources: the original \textit{$\sim${30} FPS} footage and a downsampled 10 \textit{FPS} version. Manual object counts were used as the ground-truth for quantity assessment. This comparative analysis focuses on three aspects: the accuracy of MV/VRU position and count, overall speed distribution, and individual speed profiles.

\begin{figure}[htb]
	\centering
	\includegraphics[width=0.7\textwidth]{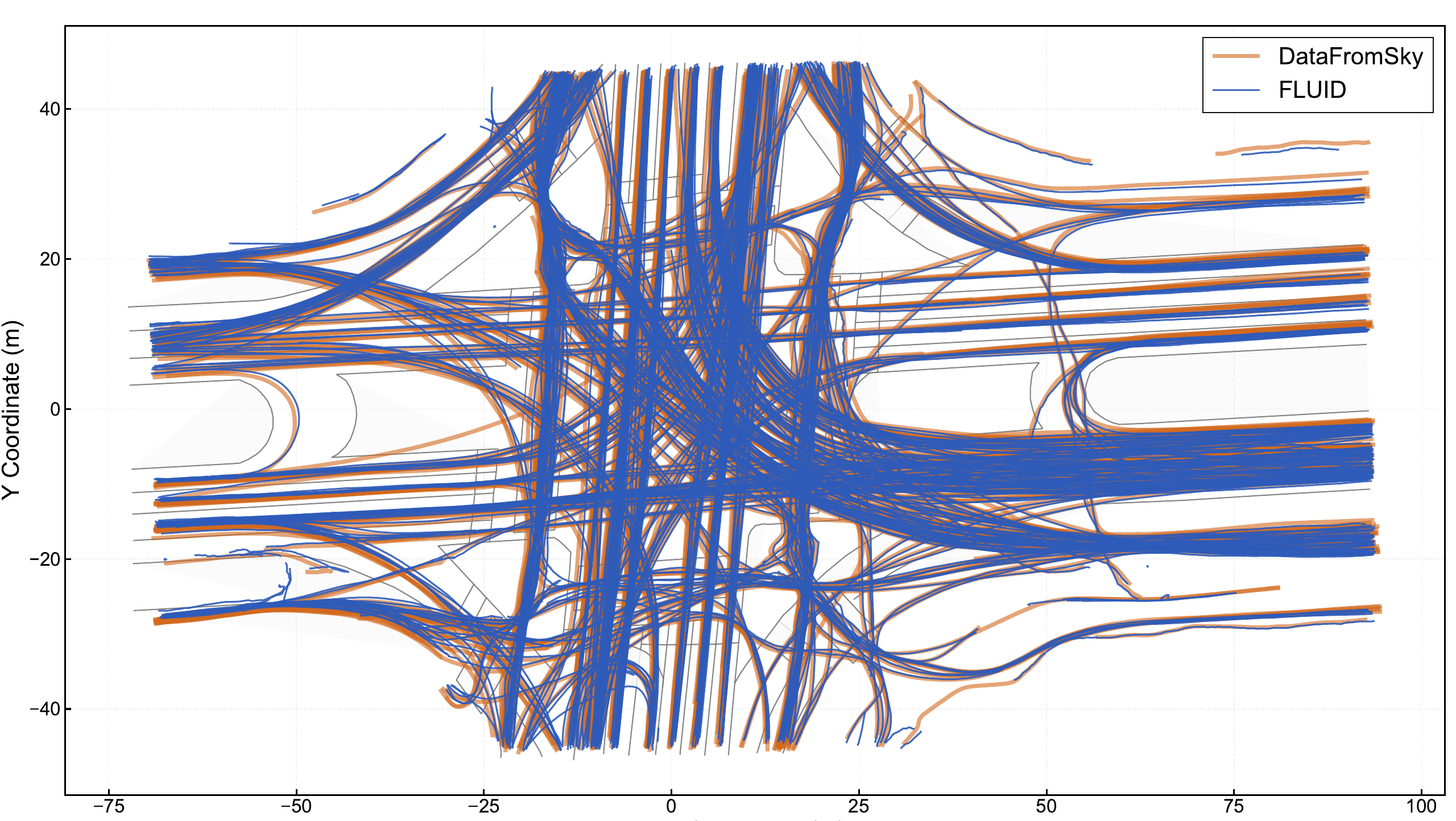}
	\caption{Position comparison (the coordinate of trajectories)} 
	\label{fig:type}
\end{figure}
	
\begin{table}[htbp]
	\centering
	\caption{Comparison of MV and VRU counts across different sources}
	\begin{tabular}{lcccc}
		\hline
		\multirow{2}{*}{Type} & \multicolumn{3}{c}{Source} & {Error} \\
		\cline{2-4}
		& {FLUID} & {DFS} & {Ground Truth} & {(\%)} \\
		\hline
		MV  & 288 & 267 & 281 & \textbf{+2.5}/-5.0 \\
		VRU & 206 & 184 & 196 & \textbf{+5.1}/-6.1 \\
		Total & 494 & 451 & 477 & \textbf{+3.6}/-5.4 \\
		\hline 
	\end{tabular}
	\label{tab:source_compare}
\end{table}

\paragraph{Position and Count.} Figure \ref{fig:type} presents the aligned trajectories positions. For consistency, all object classes were grouped into MV and VRU categories. The spatial comparison reveals that FLUID-extracted trajectories closely correspond with the DFS output. This positional accuracy is further corroborated by the RTS-GNSS ground truth: the Hausdorff distance between the reference and extracted trajectories for the test vehicle ranges $0\sim{0.97} m$, with errors under $0.3 m$ on straight segments. Table \ref{tab:source_compare} reveals object counts. DFS exhibited a $5\sim{6}\%$ miss rate against the ground-truth, a consequence of its policy to discard stationary and short trajectories. While this may improve tracking precision, it sacrifices recall. In contrast, FLUID achieved near-zero missed detections and a low ID switch rate of $2\sim{5} \%$, demonstrating performance comparable or superior to the DFS benchmark in comprehensive object detection.

\begin{figure}[htb]
	\centering
	\includegraphics[width=\textwidth]{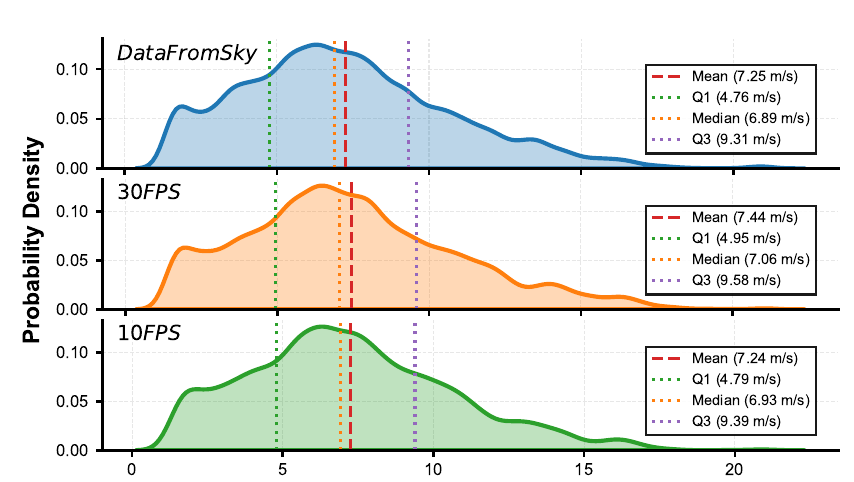}
	\caption{Speed distribution of MVs of different methods}
	\label{fig:speed_distrib}
\end{figure}

\paragraph{Speed Distribution.} Figure \ref{fig:speed_distrib} presents the overall MV speed distributions from DFS and FLUID. A notable finding is that the speed profile from the 10 \textit{FPS} video appears more stable. We hypothesize that the lower frame rate mitigates detection jitter from bounding boxes, suggesting that processing at very high frame rates may, counterintuitively, complicate the velocity post-processing stage.

\begin{figure}[htb]
	\centering
	\includegraphics[width=\textwidth]{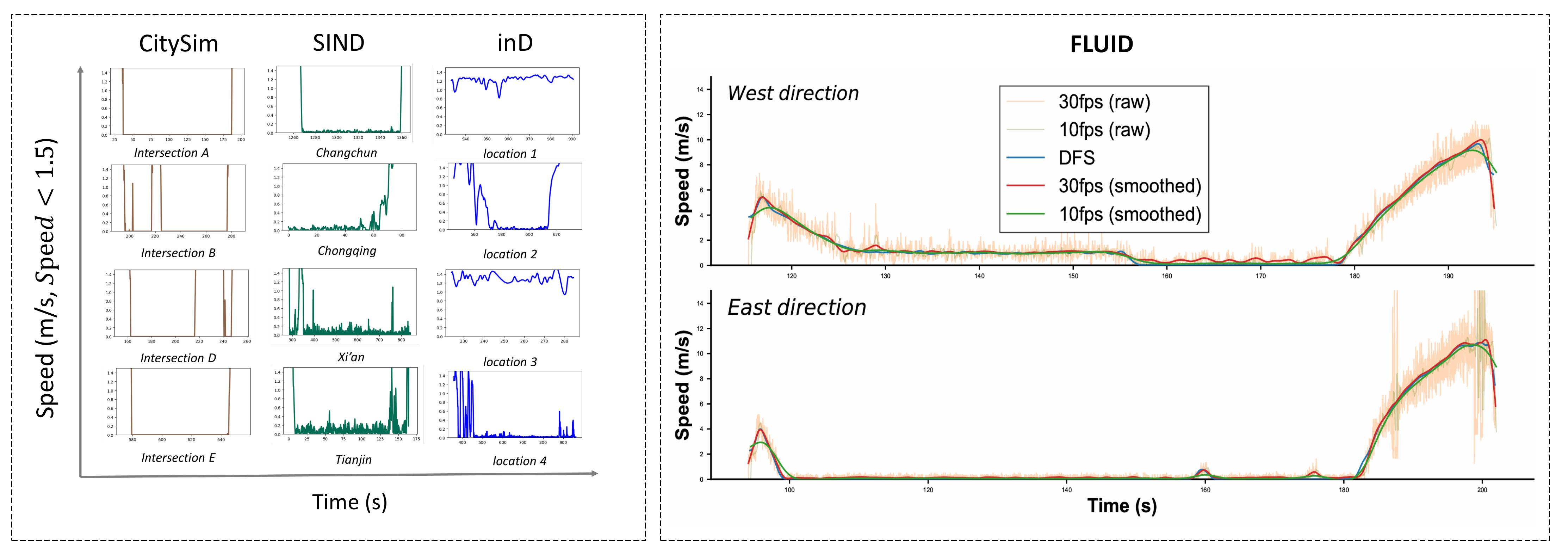}
	\caption{Comparison of processed results of speed} 
	\label{fig:speed_smooth}
\end{figure}

\paragraph{Individual Speed Profiles.} This observation is reinforced by examining individual speed profiles (the two MVs with the longest trajectories in the FLUID scenario). In contrast, the left panel of Figure \ref{fig:speed_smooth} illustrates that existing datasets employ inconsistent smoothing strategies. For instance, CitySim thresholds all near-zero speeds to zero, whereas SIND and inD apply this only to persistently stationary objects, leaving transient stops untreated. The variation in smoothing granularities across scenes—as seen in datasets like SIND and inD—is substantial and could lead to analytical biases. Conversely, FLUID enhances transparency by providing both raw and smoothed data (Figure \ref{fig:speed_smooth}, right). The analysis again demonstrates that the 10 \textit{FPS} processing yields accurate, stable speed profiles that align well with the DFS results, thus validating our methodological choices.

\subsection{Scene Significance}

A central application of the FLUID dataset is traffic conflict analysis. Defined as hazardous traffic interactions \cite{klebelsberg1964derzeitiger}, conflicts serve as a proxy for collisions, and their link to accident risk can be quantified using Surrogate Safety Measures (SSMs) \cite{wang2021review}. Due to the lack of standardized SSM thresholds across different scenarios, we developed a new conflict quantification process based on a comprehensive literature review. This process includes conflict extraction, classification, and the identification of associated TPs.

To maintain a consistent comparative basis, our analysis is confined to MVs. We selected Time-to-Collision (TTC) as the primary predictive SSM, given its robustness after standardizing the temporal resolution of velocity data across datasets. Potential conflicts are initially identified using the minimum TTC (minTTC) observed between any two trajectories \cite{wang2021review}. Subsequently, to improve precision, we introduce our Dynamic Gap Time (DGT) method for post-validation, which effectively filters out kinematically plausible but physically impossible conflicts (e.g., those separated by infrastructure). Drawing from a review of established practices \cite{wang2025conflict}, we define a conflict event using the thresholds: ${0s}\le{DGT}\le{4.0s}, {0s}\le{TTC}\le{2.0s}$.

\begin{figure}[htb]
	\centering
	\includegraphics[width=0.9\textwidth]{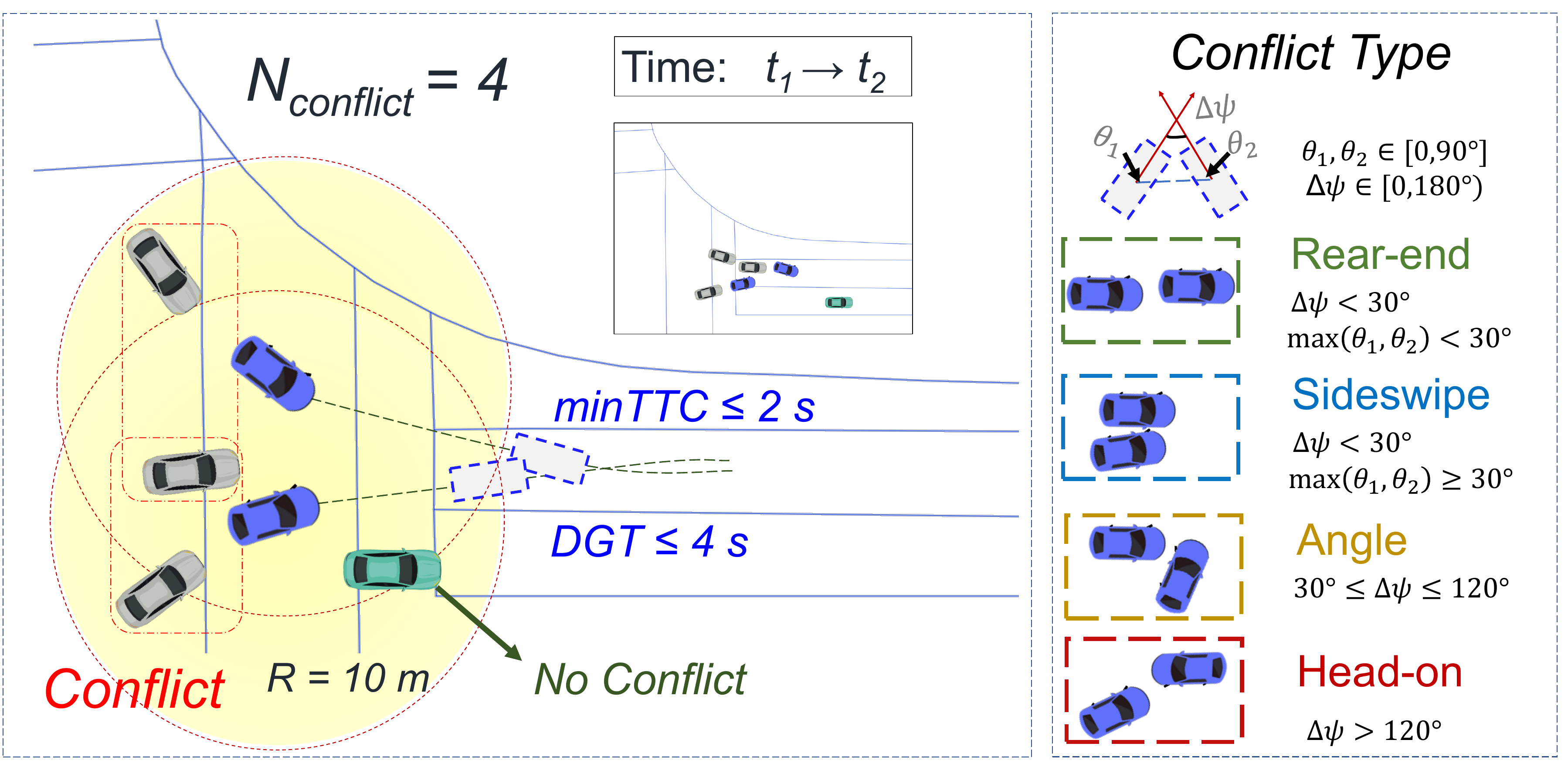}
	\caption{Traffic conflicts identification and classification} 
	\label{fig:process}
\end{figure}

The conflict angle, $\Delta{\psi}$, is a critical determinant of potential accident severity. While recent research widely adopts angle-based classification for conflicts and collisions, existing methodologies often suffer from ambiguous criteria \cite{park2025micro, anisha2023automated}, overly complex rules requiring bounding box overlap analysis \cite{wu2020automated}, or incomplete taxonomies \cite{zhang2024real}. To address these limitations, we adopt the comprehensive definition \cite{feng2023dense}, categorizing conflicts into four distinct types: rear-end, sideswipe, angle, and head-on, as depicted in Figure \ref{fig:process}.

Figure \ref{fig:process} illustrates the process of identifying associated conflicting objects . At instant $t_1$, the two blue vehicles constitute a primary conflict pair. All other vehicles within a 10-meter radius of this pair—the two gray vehicles—are designated as \textit{associated objects}, which will subsequently engage in new conflicts at a later time ($t_2$). The green vehicle, being outside this radius, is excluded. Associated objects can be repeatedly identified if they are proximate to multiple conflict TPs.

Table \ref{tab:overall} shows that FLUID's conflict quantification demonstrates clear advantages over other datasets. A key differentiator is the traffic composition surrounding conflict events. Our analysis reveals that VRUs constitute 35.4\% of all agents within a 10-meter radius of a conflict pair in FLUID. This proportion is substantially higher than that in SIND (7.2\%), inD (23.7\%), and INTERACTION (4.4\%), indicating that FLUID provides a unique environment for studying MV interactions in the presence of VRUs, which impacts driver decision-making.

\subsection{Behavioral Richness}

\begin{figure}[t]
	\centering
	\includegraphics[width=0.7\textwidth]{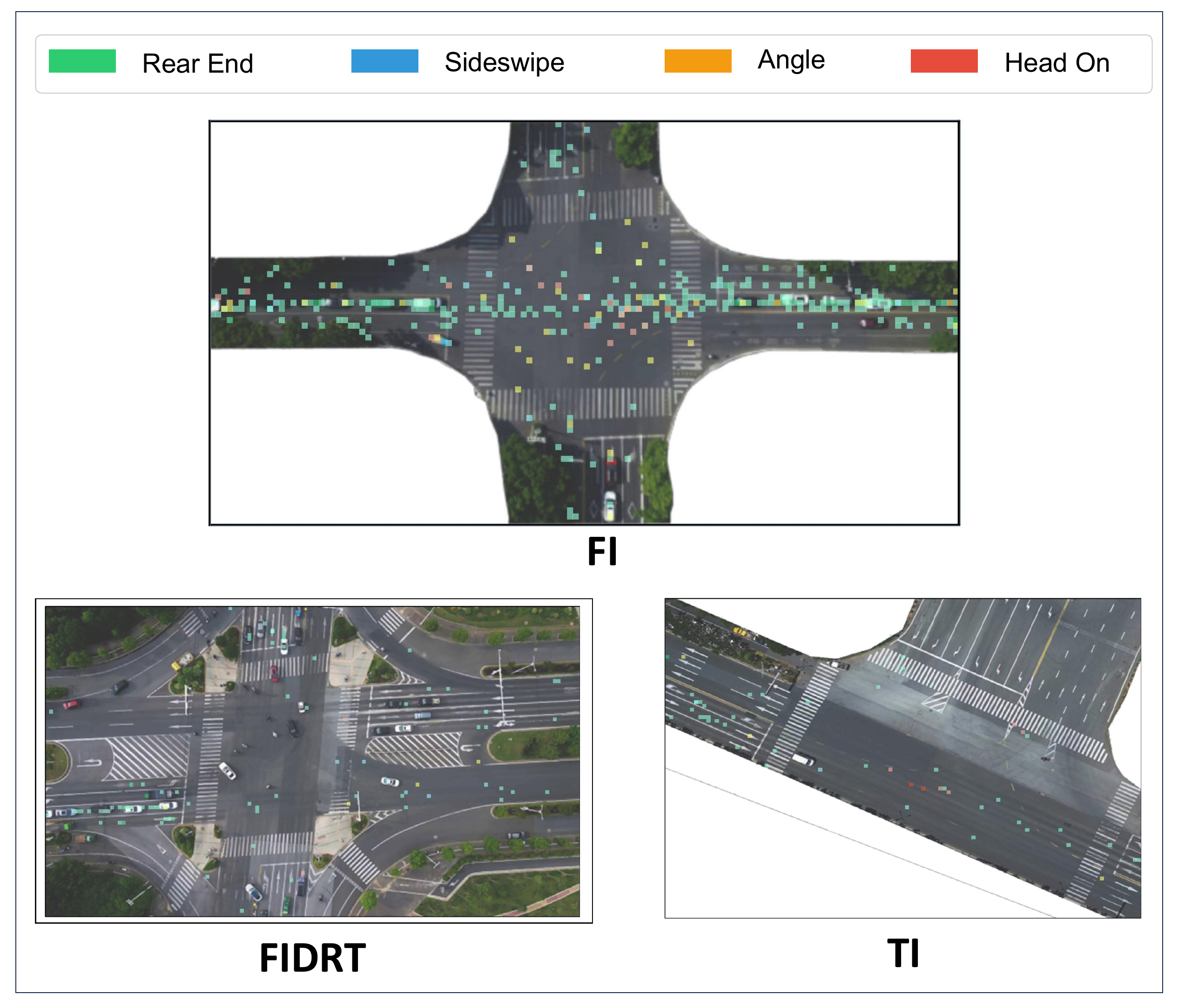}
	\caption{Heatmaps of spatial density for different conflict types} 
	\label{fig:conflictDistrib}
\end{figure}

\begin{figure}[t]
	\centering
	\includegraphics[width=0.5\textwidth]{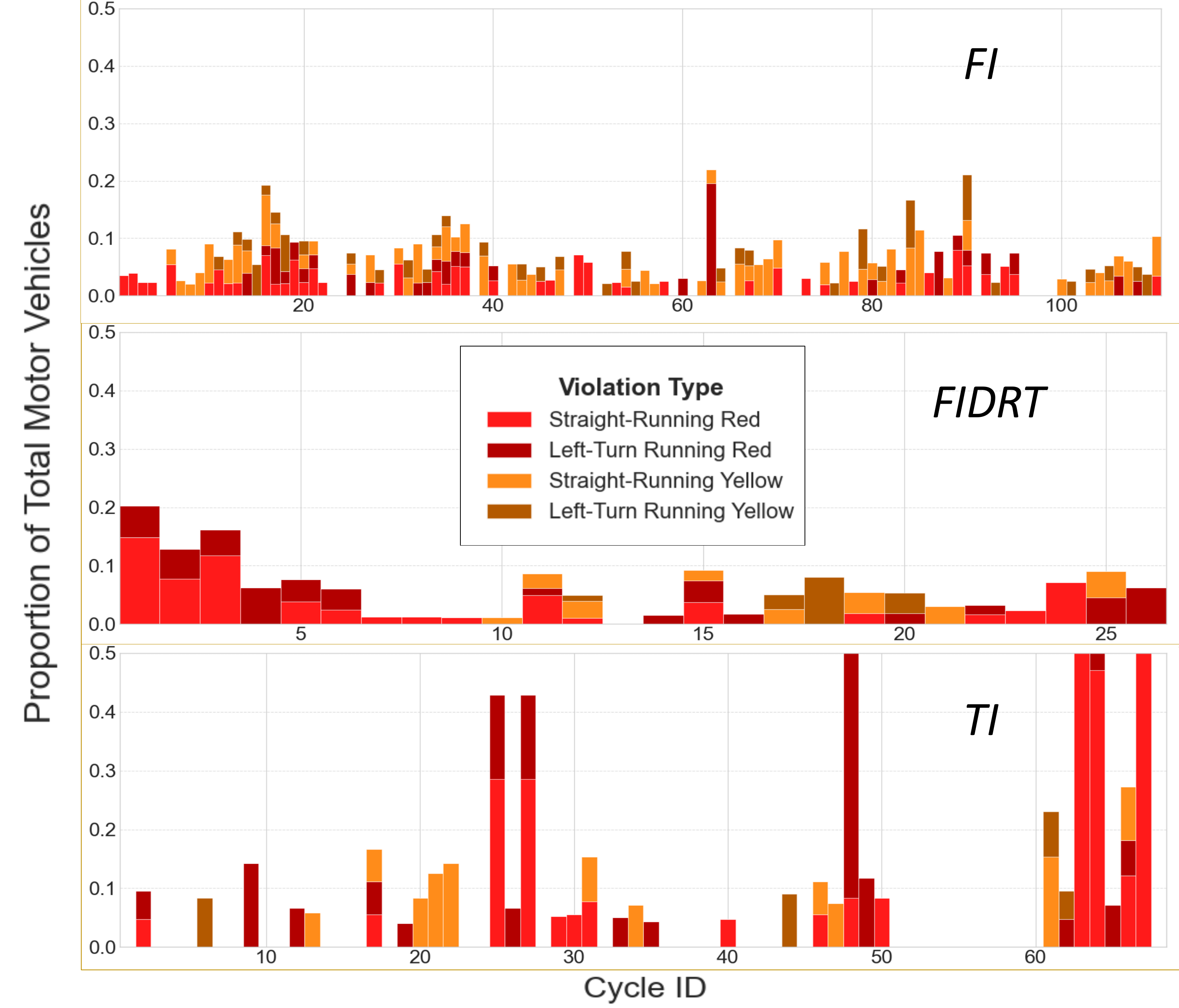}
	\caption{Per-cycle violation rates for straight and left-turn movements, calculated as a proportion of total MVs} 
	\label{fig:violationSitu}
\end{figure}

In FLUID, conflict and violation annotations exhibit rich spatiotemporal behavioral characteristics. The $1m\times{1m}$ grid is used for discrete division, enabling finer conflict identification. Subsequently, the previously classified traffic types are clustered according to the grid. Figure \ref{fig:conflictDistrib} showed that different scenarios exhibit diverse traffic conflicts type-density distributions. Figure \ref{fig:violationSitu} shows that the violation rates for each signal cycle (non-consecutive videos, 110 for FI, 26 for FIDRT, and 68 for TI) vary. The calculated violation rates may be higher during cycles with lower traffic volume.

\section*{Usage Notes}\label{usage-notes}
\addcontentsline{toc}{section}{Usage Notes}

\begin{figure}[htb]
	\centering
	\includegraphics[width=0.9\textwidth]{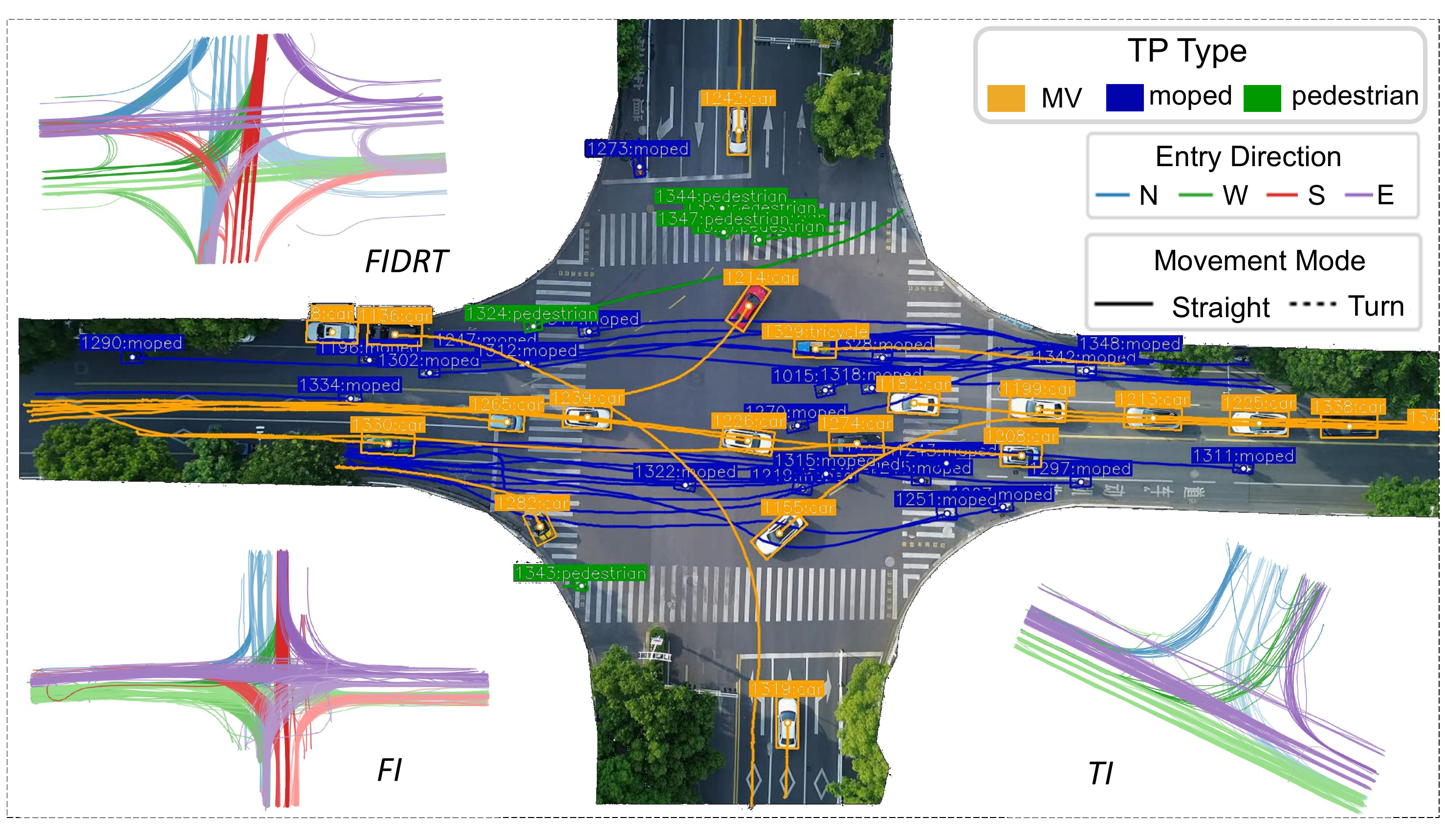}
	\caption{Trajectory visualization and turn annotation result} 
	\label{fig:turns}
\end{figure}

\begin{figure}[htb]
	\centering
	\includegraphics[width=0.9\textwidth]{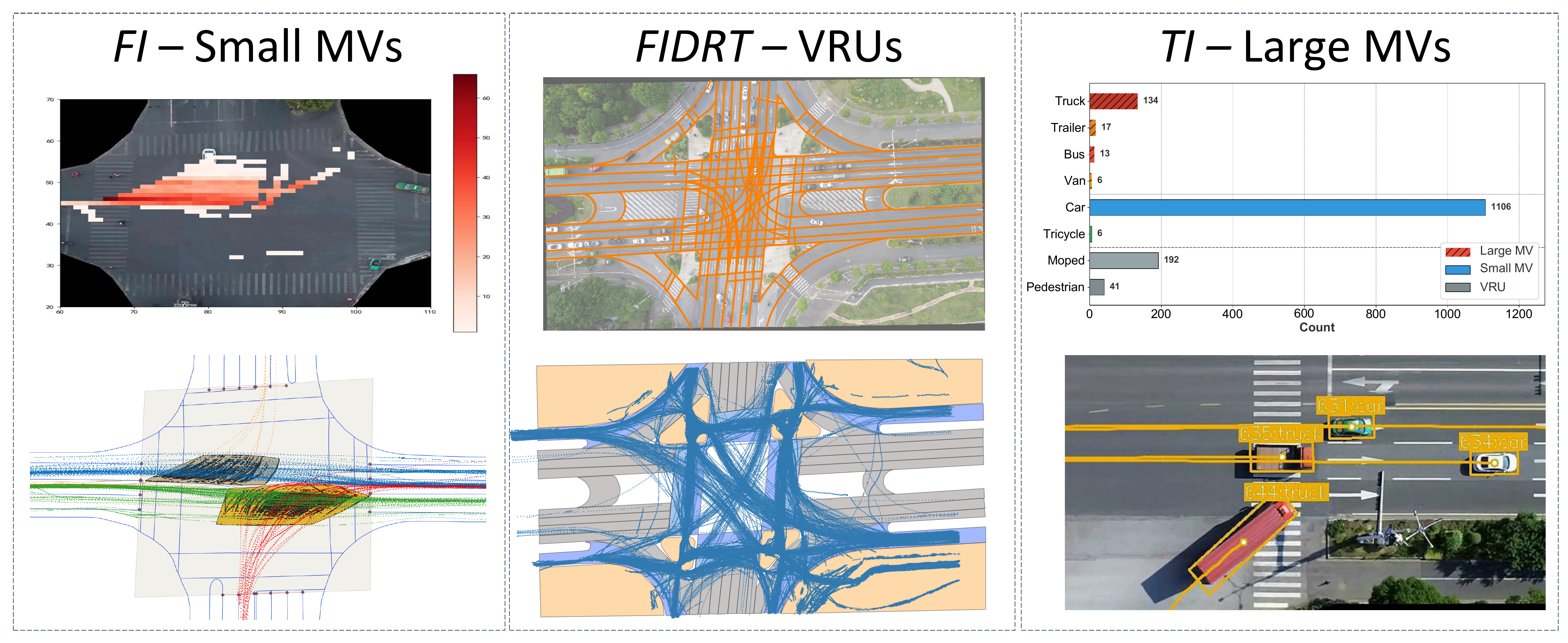}
	\caption{Unique application prospects of FLUID} 
	\label{fig:usage}
\end{figure}

Benefiting from the fine-grained details of our dataset, we can categorize distinct traffic behavior patterns through turn labels, as illustrated in Figure \ref{fig:turns}. Beyond basic traffic flow analysis, the high precision of FLUID facilitates multi-domain research, including human preference mining, traffic behavior modeling, and autonomous driving. Figure \ref{fig:usage} showcases three representative cases that highlight the dataset's unique potential in these specialized research contexts.

\paragraph{Analysis of Passing and Yielding Behaviors.}
Leveraging the standardized geometric layout, high density of passenger vehicles, and integrated traffic signal data in the FI scenario, FLUID provides a robust foundation for analyzing complex interaction behaviors. By calculating the convex hull of conflicting trajectories, we can accurately delineate conflict zones and analyze the temporal sequences of vehicles entering and exiting both these zones and the intersection boundaries. This methodology enables a refined quantification of passing and yielding dynamics for specific maneuvers. For instance, among the through-left interaction events identified during green phases across the FI dataset, we recorded 502 instances of left-turn yielding, 415 instances of through-moving yielding, and 5 instances where no clear yielding behavior was discernible (Figure \ref{fig:usage}, Left).

\paragraph{Spatiotemporal Violation Analysis of VRUs.}
While previous research\cite{XU2024129994} (conducted on a scenario similar to FIDRT) was often constrained by limited tracking precision and necessitated grid-based spatial partitioning, the FLUID dataset provides individual-level trajectories of VRUs integrated with precise Lanelet2 semantic maps. Moreover, the significantly higher VRU arrival rate in FLUID, compared to existing benchmarks, allows for a more granular analysis of spatial occupancy. These features facilitate a deeper understanding of VRUs' spatiotemporal violation intentions and movement patterns (Figure \ref{fig:usage}, Middle).

\paragraph{Optimization of Large Vehicle Detection}
In the TI scenario, large vehicles constitute over 15\% of the traffic, where their diverse dimensions present substantial challenges for accurate detection. By providing both raw data and bounding box labels, we offer significant optimization potential for large-scale target detection in complex environments. Ultimately, this provides a contribution of similar value to pNEUMA Vision \cite{kim2023visual} for multi-object tracking and detection research (Figure \ref{fig:usage}, Right).

In addition to these unique capabilities, our dataset also serves as a high-quality resource for broader applications in traffic engineering, such as:
\begin{itemize}
	\item \textit{Driving Decision Modeling}: Analyze the impact of interconnected information on driving decisions and traffic operations in interactive dilemmas involving two vehicles encountering conflicting directions at an intersection \cite{yang2025does}.
	\item \textit{Intersection Operation Quantification}: Evaluate the safety and efficiency characteristics of different intersection control strategies \cite{pu2025drone}.
	\item \textit{Conflict Correlation Analysis}: Quantify conflict severity using SSMs, and analyze the relationship between this severity and other kinematic parameters \cite{shen2024analysis}.
	\item \textit{Trajectory Generation}: Learn from human mobility patterns to generate human-like and socially-inspired behaviors and movements \cite{wang2025multiagent}.
\end{itemize}

\section*{Data Availability}\label{data-availability}
\addcontentsline{toc}{section}{Data Availability}

The full dataset is available at figshare \cite{Chen2025}.
The DOI to data record is \href{https://doi.org/10.6084/m9.figshare.29974954}{10.6084/m9.figshare.29974954}.

\section*{Code Availability}\label{code-availability}
\addcontentsline{toc}{section}{Code Availability}

Regarding raw data processing, the object detection and tracking code itself is based on the open-source YOLOv8 and SparseTrack. Therefore, the entire code is not repeated. The methods for processing, verification and visualization of the FLUID dataset is publicly available on GitHub: \href{https://github.com/sysu19351014/FLUID/tree/main}{https://github.com/sysu19351014/FLUID}.

\nolinenumbers


\section*{Acknowledgements}\label{acknowledgements}
\addcontentsline{toc}{section}{Acknowledgements}
We are grateful to the Joint Research and Development Laboratory of Smart Policing in Xuancheng Public Security (a joint initiative between Xuancheng Public Security Bureau and Sun Yat-sen University) for their essential contribution to this work, specifically in facilitating the selection of the investigation's spatiotemporal scope and data acquisition. We thank Junfeng Li and Kaiyuan Yang for their assistance in vector map processing, as well as the anonymous undergraduate students from the School of Intelligent System Engineering at SYSU for their help with data integration. The research was also supported by the National Key Research and Development Program of China (No. 2023YFB4301900), the National Natural Science Foundation of China (No. U21B2090), the Shenzhen Science and Technology Program (No.JCYJ20240813151445059), and the Science and Technology Planning Project of Guangdong Province (No. 2023B1212060029).

\section*{Author Contributions}\label{author-contribution}
\addcontentsline{toc}{section}{Author Contributions}

Conceptualization and Methodology: Y.C., Z.W., G.Z., H.Z.; Investigation: Y.C., Z.W., L.X.; Data Curation and Formal Analysis: Y.C., G.Z.; Validation and Visualization: Y.C., L.X., H.T., Z.W.; Writing – Original Draft Preparation: Y.C.; Writing – Review \& Editing: Z.W, X.W., Z.H.; Project Administration and Funding Acquisition: H.Z., Z.H.  All authors have reviewed and agreed to the manuscript.

\section*{Competing Interests}\label{competing-interests}
\addcontentsline{toc}{section}{Competing Interests}

The authors declare no competing interests.

%
%
%
%

\end{document}